\documentclass{article}
\usepackage{PRIMEarxiv}
\errorcontextlines\maxdimen

\usepackage[T1]{fontenc}
\usepackage[utf8]{inputenc}
\usepackage{newtxtext}
\usepackage{microtype}
\usepackage[dvipsnames,table]{xcolor}
\usepackage{graphicx}
\usepackage{booktabs}
\usepackage{multirow}
\usepackage{array}
\usepackage{tabularx}
\usepackage{makecell}
\usepackage{amsmath,amssymb,amsfonts,mathtools}
\usepackage{newtxmath}
\usepackage{bm}
\usepackage{enumitem}
\usepackage{caption}
\usepackage{subcaption}
\usepackage[section]{placeins}
\usepackage{float}
\usepackage{fancyhdr}
\usepackage{titlesec}
\usepackage[most]{tcolorbox}
\tcbuselibrary{breakable,skins}
\usepackage{tikz}
\usepackage[numbers,sort&compress]{natbib}
\usepackage{xurl}
\usepackage{hyperref}

\hypersetup{
  colorlinks, citecolor=blue, linkcolor=red, urlcolor=green,
  pdftitle={See, Hypothesize, Validate: Multimodal Agentic Framework for Discovering Governing PDEs},
  pdfauthor={Sarang Manoj Pekhale, Amartya Roy, Rajat Sarkar, Souvik Chakraborty}
}

\setlist[itemize]{leftmargin=1.35em,itemsep=2pt,topsep=4pt}
\setlist[enumerate]{leftmargin=1.55em,itemsep=2pt,topsep=4pt}

\definecolor{MageNavy}{HTML}{28566A}
\definecolor{MageRule}{HTML}{4F7C8D}
\definecolor{MageStripe}{HTML}{F2F7F9}
\definecolor{MageAccent}{HTML}{E7F0FA}
\definecolor{MageGood}{HTML}{E1F1E8}
\definecolor{MageWarn}{HTML}{FFF1D2}
\definecolor{MageBad}{HTML}{F8E2DF}
\definecolor{MageSpurious}{HTML}{F6D68A}
\definecolor{MageNoiseExcellent}{HTML}{CFEAE6}
\definecolor{MageNoiseGood}{HTML}{DDEFD8}
\definecolor{MageNoiseModerate}{HTML}{F7E8A6}
\definecolor{MageNoiseHigh}{HTML}{F5C995}
\definecolor{MageNoiseCritical}{HTML}{F3C1BD}
\definecolor{MageNoiseClean}{HTML}{EAF6F3}
\arrayrulecolor{MageRule}
\newcommand{\mageTableRows}{\rowcolors{2}{MageStripe}{white}}
\newcommand{\mageTableHead}{\rowcolor{white}}

\titleformat{\section}
  {\Large\bfseries}{\thesection}{0.7em}{}
\titleformat{\subsection}
  {\large\bfseries}{\thesubsection}{0.65em}{}
\titleformat{\subsubsection}
  {\normalsize\bfseries}{\thesubsubsection}{0.6em}{}
\titlespacing*{\section}{0pt}{2.2ex plus .8ex minus .2ex}{0.8ex}
\titlespacing*{\subsection}{0pt}{1.8ex plus .6ex minus .2ex}{0.6ex}
\titlespacing*{\subsubsection}{0pt}{1.15ex plus .4ex minus .2ex}{0.4ex}

\title{See, Hypothesize, Validate: Multimodal Agentic Framework\\
for Discovering Governing PDEs}

\author{
Sarang Manoj Pekhale\thanks{Equal contribution} \\
Department of Applied Mechanics \\
Indian Institute of Technology Delhi, India \\
\texttt{amz248506@am.iitd.ac.in} \\
\And
Amartya Roy\textsuperscript{*} \\
The School of Interdisciplinary Research \\
Indian Institute of Technology Delhi, India \\
Robert Bosch GmbH, India \\
\texttt{srz248670@iitd.ac.in} \\
\And
Rajat Sarkar \\
Department of Applied Mechanics \\
Indian Institute of Technology Delhi, India \\
TCS Research, India \\
\texttt{rajat.sarkar1@tcs.com} \\
\And
Souvik Chakraborty \\
Department of Applied Mechanics \\
Indian Institute of Technology Delhi, India \\
Yardi School of Artificial Intelligence (ScAI) \\
Indian Institute of Technology Delhi, India \\
\texttt{souvik@am.iitd.ac.in}
}

\begin{document}
\maketitle

\begin{abstract}
Discovering governing partial differential equations (PDEs) from observational data remains a core challenge across the sciences. Existing sparse-regression, symbolic-regression, and LLM-based approaches can be constrained by predefined libraries, noise sensitivity, hallucination, or limited iterative refinement. We introduce \textbf{MAGE} (\textbf{M}ultimodal \textbf{A}gentic \textbf{G}overning \textbf{E}quation Discovery), an agentic framework that organizes PDE discovery as a \textit{confidence governed hypothesis validation loop} inspired by the scientific cycle of observation, hypothesis, and falsification. Four role-specialized agents collaborate: a \textit{Differential Observer} computing derivatives and diagnostic visualizations; a VLM-powered \textit{Phenomenology Extractor} distilling qualitative cues from multimodal diagnostics; an LLM-driven \textit{Governing Law Synthesizer} proposing candidates without a predefined library; and an \textit{Equation Arbiter} fitting coefficients and assigning confidence scores. Discovery iterates until the top candidate clears a user-specified threshold, providing a structured process with an explicit accept-reject protocol. On the evaluated canonical PDE suite, MAGE obtains \textbf{8/8} exact structural recovery and the lowest coefficient error among the compared methods on \textbf{7/8} systems, with improvements of up to \textbf{4 orders of magnitude} and a geometric-mean improvement of approximately \textbf{3 orders of magnitude}. The pipeline also recovers the expected operators in two complex geometries and, on one laboratory sensor record, selects a cubic restoring-force model with held-out $R^2=0.98538$. These results support further study of structured agentic reasoning for library-free governing-law discovery, while broader generalization remains to be evaluated.
\end{abstract}

\section{Introduction}
When a scientist encounters an unknown physical system, discovery often begins with observation: reading contour fields, interpreting spectral signatures, and tracing how disturbances propagate across space and time. These qualitative patterns can inform first-principles reasoning about governing partial differential equations (PDEs), but the process remains slow and expert dependent. Many complex systems across neuroscience, materials science, chemistry, and biology generate rich observational data without a settled governing PDE. Recent advances in machine learning offer tools for extracting physical patterns from measurements and imagery and for assisting parts of the observation-to-hypothesis cycle.

Existing data driven equation discovery frameworks broadly fall into four categories: library based methods, symbolic regression, LLM based generation, and, more recently, agentic frameworks. As detailed in Related Work, each is constrained in a distinct way: fixed candidate libraries, combinatorial search over open form expressions, hallucination and prompt sensitivity, or reliance on a single general purpose model without task specialization \cite{brunton2016discovering, schmidt2009distilling, du2024llm4ed, xia2026srscientist}.

Despite differences in implementation, these paradigms often emphasize a direct mapping from data to equations, with less explicit support for iterative interaction with observational evidence. We instead consider a cycle of pattern extraction, hypothesis formation, quantitative validation, and refinement. To this end, we propose MAGE, a multimodal agentic framework that operationalizes PDE discovery as a confidence governed hypothesis validation loop of four role specialized agents (Section~\ref{sec:methodology}), inspired by the scientific cycle of observation, hypothesis, and falsification. Discovery iterates until the top ranked candidate exceeds a user specified confidence threshold. Although the underlying LLMs and VLMs can hallucinate in isolation, conditioning successive stages on multimodal evidence and numerical feedback is intended to reduce, rather than eliminate, physically inconsistent candidates without a predefined symbolic library.

Our contributions are as follows.

\begin{itemize}
\item \textbf{A latent variable view of agentic discovery.} We cast PDE discovery as inference over a hierarchy of intermediate latent states, namely numerical diagnostics, semantic physical abstractions, a slate of candidate equations, and validation metrics. This factorization (Section~\ref{sec:latent}) assigns each conditional factor to one role specialized agent, replacing a single broad combinatorial search with a sequence of smaller conditioned reasoning steps.

\item \textbf{Visual evidence as an explicit conditioning signal.} Prior LLM based discovery methods condition on numerical arrays, symbol libraries, or program templates. MAGE instead renders contour, gradient, and spectral diagnostics and uses a VLM to extract qualitative regime descriptors such as advection dominance, diffusive smoothing, and nonlinear transport, which then constrain how candidate equations are proposed.

\item \textbf{An explicit accept reject protocol.} Candidates are instantiated as executable validation programs and scored numerically against the observed data. The entire slate is rejected and regenerated when confidence falls below a user specified threshold, giving discovery a stopping criterion tied to numerical evidence rather than a fixed iteration budget.

\item \textbf{Evaluation across canonical, geometric, and experimental regimes.} We evaluate on a canonical PDE suite, on two complex geometries, and on a laboratory sensor record. MAGE recovers exact structure on 8/8 canonical systems and attains the lowest coefficient error on 7/8, improving on the compared methods by up to four orders of magnitude.
\end{itemize}

 \section{Related Work}
Data driven PDE discovery has been extensively studied under four main paradigms: library based methods, symbolic regression, LLM based generation, and agentic frameworks.

\textbf{Library Based Approaches.} These methods formulate PDE discovery as sparse regression over a predefined library of candidate terms, from the seminal SINDy framework \cite{brunton2016discovering} to extensions improving robustness via sequential thresholding \cite{rudy2017data,schaeffer2017learning,fasel2022ensemble}, weak-form formulations that avoid explicit differentiation \cite{messenger2021weak,reinbold2020using}, automatic-differentiation based neural approaches \cite{raissi2019physics,both2021deepmod}, and Bayesian uncertainty aware sparsity \cite{more2023bayesian}. Their expressivity nevertheless depends on which terms are included in the predefined function space.

\textbf{Symbolic Regression Methods.} These methods reduce fixed-library constraints by searching over mathematical expressions. Early genetic programming approaches \cite{schmidt2009distilling,udrescu2020aifeynman,cranmer2023pysr} evolve candidates via mutation and recombination, which can make search computationally expensive as expression complexity grows. Reinforcement learning based methods instead learn policies over expression construction for greater efficiency \cite{petersen2021deep}, while grammar and structure constrained models stabilize search in PDE discovery settings \cite{du2024discover}. Symbolic regression can nevertheless remain sensitive to noise and face a rapidly growing search space in multivariate systems.

\textbf{LLM Based Approaches.} Recent methods leverage pretrained language models as priors over mathematical structures: LLM4ED generates candidates from symbol libraries with iterative refinement \cite{du2024llm4ed}, LLM-SR embeds generation within programmatic templates optimized via evolutionary search \cite{shojaee2025llmsr}, LLM4PD decomposes term selection and coefficient fitting via hierarchical prompting \cite{luo2025llm4pd}, and EqGPT shows lightweight models trained on curated PDE corpora can generate candidates without large foundation models \cite{xu2025eqgpt}. Their outputs can still depend on prompting and on the implicit or explicit symbolic vocabulary available to the model.

\textbf{Agentic Frameworks.} These approaches incorporate iterative reasoning and tool use into equation discovery pipelines: SR-Scientist equips a single LLM with code execution tools to iteratively propose and refine equations \cite{xia2026srscientist,wu2024autogen,bran2024chemcrow}, while KeplerAgent leverages physics informed priors such as symmetries and conservation laws \cite{yang2026kepleragent}. Existing demonstrations are promising, but many use a single general purpose model and have so far been evaluated mainly on algebraic or low dimensional forms rather than a broad range of complex spatiotemporal PDEs.
\section{Methodology}
\label{sec:methodology}

\subsection{Problem Formulation}
\label{sec:problem_formulation}

Given field observations $\mathcal{D}={(\mathbf{z}^{(i)},\tilde{\bm u}^{(i)})}_{i=1}^{N}$ of an unknown field $\bm u=\bm u(\mathbf{z})\in\mathbb{R}$, where $\mathbf{z}\in\mathbb{R}^{d}$ denotes the sensor locations and, when applicable, the temporal sampling instants at which measurements are acquired, and $\tilde{\bm u}$ is the observed field response, the PDE discovery task seeks an explicit governing law of the form

\begin{equation} 
\mathcal{F}\left(\bm u,\nabla \bm u,\nabla^{2} \bm u,\ldots;\boldsymbol{\theta}\right)=0,
\label{eq:pde_form}
\end{equation}

parameterized by coefficients $\boldsymbol{\theta}$ and expressed in terms of the field and its partial derivatives (spatial and temporal). We consider both time dependent and time independent systems within a single formulation. Since $\mathcal{F}$ is drawn from a combinatorial hypothesis space $\mathcal{H}$ and $\boldsymbol{\theta}$ is unknown, discovery is naturally posed as a maximum likelihood estimation problem:

\begin{equation}
(\mathcal{F}^{\ast},\boldsymbol{\theta}^{\ast}) =
\arg \max_{\mathcal{F}\in\mathcal{H},\boldsymbol{\theta}}
p\left(\mathcal{D}\mid\mathcal{F},\boldsymbol{\theta}\right).
\label{eq:mle}
\end{equation}

Direct optimization over $\mathcal{H}$ is computationally intractable due to the combinatorial growth of candidate operators and the continuous optimization over coefficient space. The core challenge lies in resolving this coupled symbolic parametric inference problem under combinatorial complexity while retaining sufficient expressivity to recover the underlying governing law.

\begin{figure}[t]
  \centering
  \includegraphics[width=\textwidth]{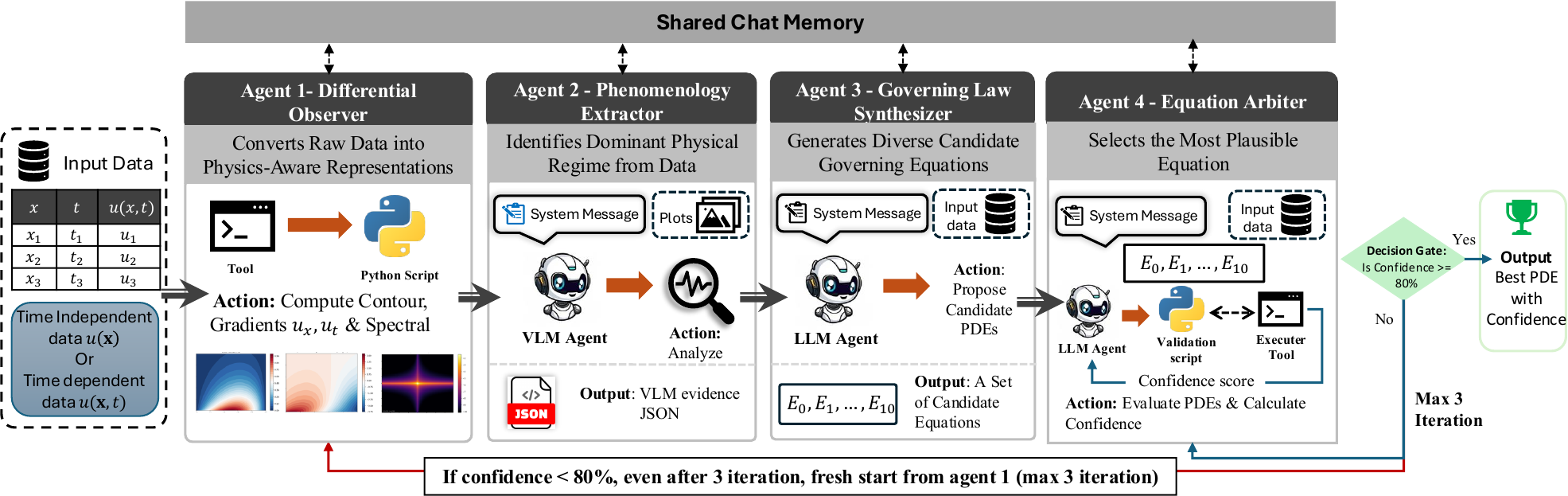}
  \caption{%
    \textbf{MAGE Framework.} Four specialized agents communicate via a \emph{Shared Chat Memory}. \textbf{Agent~1} computes gradients and spectral features from raw data. \textbf{Agent~2} uses a VLM to identify dominant physical regimes and outputs a JSON evidence report. \textbf{Agent~3} conditions an LLM on this evidence to propose $N\!=\!10$ candidate PDEs $\{E_0,\ldots,E_{9}\}$. \textbf{Agent~4} evaluates candidates symbolically; if confidence $\geq\!80\%$ the top equation is returned, otherwise the loop restarts from Agent~1 for up to three iterations.
}
  \label{fig:framework}
\end{figure}

\subsection{Agentic Latent Variable Decomposition}
\label{sec:latent}

To resolve the combinatorial reasoning bottleneck in Eq.~\eqref{eq:mle}, we reformulate PDE discovery as a structured latent variable inference problem. Rather than directly optimizing over the joint symbolic parametric space $(\mathcal{F},\boldsymbol{\theta})$, we introduce a hierarchy of intermediate latent variables that progressively transform raw field observations into validated governing law hypothesis. Specifically, we define $\mathcal{Z}_1$ as numerical diagnostic features extracted from the observed field, including differential signatures, spectral characteristics, and geometric contours; $\mathcal{Z}_2$ as semantic physical abstractions inferred from these diagnostics (e.g., advection dominance, diffusive smoothing, nonlinear transport); $\mathcal{Z}_3=\{\mathcal{F}_k\}_{k=1}^{K}$ as a slate of candidate governing equations; $\{\mathcal{C}_k\}_{k=1}^{K}$ as executable validation programs instantiated for each candidate; and $\mathcal{Z}_4=\{\mathcal{L}_k\}_{k=1}^{K}$ as the resulting evaluation metrics quantifying numerical consistency with the observed data. Formally, we factorize the posterior over governing laws as:

\begin{equation}
\label{eq:factorization}
\begin{split}
p(\mathcal{F},\boldsymbol{\theta}\mid\mathcal{D})
&=
\sum_{\substack{
\mathcal{Z}_1,\mathcal{Z}_2,\mathcal{Z}_3,\mathcal{Z}_4,\\[-1pt]
\{\mathcal{C}_k\}
}}
p\left(
\begin{gathered}
\mathcal{F},\boldsymbol{\theta},
\mathcal{Z}_1,\mathcal{Z}_2,\\[-1pt]
\mathcal{Z}_3,\mathcal{Z}_4,
\{\mathcal{C}_k\}\mid\mathcal{D}
\end{gathered}
\right)
\\
&=
\sum_{\substack{
\mathcal{Z}_1,\mathcal{Z}_2,\mathcal{Z}_3,\mathcal{Z}_4,\\[-1pt]
\{\mathcal{C}_k\}
}}
\underbrace{
p(\mathcal{Z}_1\mid\mathcal{D})
}_{\substack{
\text{Differential}\\[-1pt]
\text{Observer}
}}
\\
&\quad
\underbrace{
p(\mathcal{Z}_2\mid\mathcal{Z}_1)
}_{\substack{
\text{Phenomenology}\\[-1pt]
\text{Extractor}
}}
\;
\underbrace{
p(\mathcal{Z}_3\mid\mathcal{Z}_2,\mathcal{D})
}_{\substack{
\text{Governing Law}\\[-1pt]
\text{Synthesizer}
}}
\\
&\quad\cdot
\underbrace{
p(\{\mathcal{C}_k\}\mid\mathcal{Z}_3)\,
p(\mathcal{Z}_4,\boldsymbol{\theta},\mathcal{F}
\mid\{\mathcal{C}_k\},\mathcal{D})
}_{\substack{
\text{Equation}\\[-1pt]
\text{Arbiter}
}},
\end{split}
\end{equation}

where marginalization is performed over all intermediate latent states. This factorization organizes symbolic regression into sequential conditional inference stages, each conditioned on upstream evidence. In practice, this recasts a broad combinatorial search as a structured sequence of smaller reasoning tasks.

We operationalize this decomposition through MAGE, an agentic hypothesis validation loop that mirrors the scientific cycle of observation, abstraction, hypothesis generation, and falsification (Fig.~\ref{fig:framework}). Each factor in Eq.~\eqref{eq:factorization} is instantiated by a role specialized agent with a fixed input output contract and deterministic routing rules: the Differential Observer (Agent 1) extracts diagnostics from raw measurements, the Phenomenology Extractor (Agent 2) maps these into semantic physical priors, the Governing Law Synthesizer (Agent 3) proposes candidate PDEs conditioned on both, and the Equation Arbiter (Agent 4) generates validation routines and numerically adjudicates consistency (each detailed in Agent Realization below).
A candidate is accepted only if it satisfies numerical validation criteria; otherwise the entire hypothesis slate is rejected and a new inference round begins. This confidence governed rejection is designed to reduce spurious structures through downstream numerical verification. The decomposition offers three practical advantages: it stages the symbolic search, enables reasoning strategies tailored to each subtask, and exposes intermediate states that make the derivation of a selected equation more inspectable.

\subsection{Agent Realization}

We now detail the computational realization of the latent inference stages in Eq.~\eqref{eq:factorization}. Each stage is instantiated as a task specific agent with a structured input output interface and domain specific execution tools. Agents operate sequentially under delayed terminal feedback, where numerical validation is performed only after complete candidate synthesis and the resulting reward signal is propagated backward to guide iterative refinement. 

\paragraph{Agent 1: Differential Observer.}

The Differential Observer instantiates the inference term $p(\mathcal{Z}_1\mid\mathcal{D})$ as a deterministic, code backed transformation $f_{\mathrm{feature}}:\mathcal{D}\mapsto\mathcal{Z}_1$, which maps raw field measurements and coordinate metadata into diagnostic numerical features. Its probabilistic realization is therefore a point mass,

\begin{equation}
    p(\mathcal{Z}_1\mid\mathcal{D}) = \delta\!\left(\mathcal{Z}_1-f_{\mathrm{feature}}(\mathcal{D})\right),
\end{equation}

where $\delta(\cdot)$ denotes the Dirac delta and $ \mathcal{Z}_1= \left\{ \nabla u,\, \nabla^2u,\, \hat{u}(\omega),\, \mathrm{contours}(u) \right\}.$

Holding this transformation deterministic removes one source of run-to-run variability, although it does not remove derivative-estimation error in noisy observations. For time dependent fields, the Observer produces eight diagnostics: contour $u$, derivatives $u_x$ and $u_t$, power spectra $E(k)$, extrema trajectories from zero crossings of $u_x$, the dispersion spectrum $\log_{10}|\widehat{u}(k,\omega)|^2$, spectral energy evolution $E(k,t)$, and spectral entropy. For time independent fields, output adapts to sampling topology: 2D point clouds yield scatter, contour, and triangulated surface plots, while 3D fields use orthogonal projections (refer to Appendix). This separation supplies a common numerical evidence layer for downstream reasoning.

\subsubsection{Wavelet Denoising}
\label{sec:appendix_wavelet_denoising}
The Differential Observer (Agent 1) uses the following noise injection and denoising
procedure before constructing derivative and visual evidence from structured
grid data.

For a clean structured field $u$ and a requested noise level $\rho$ percent,
the loader forms
\begin{equation}
  u^{\mathrm{noisy}} = u + \frac{\rho}{100}\,\sigma_u\,\eta,
  \qquad
  \eta \sim \mathcal{N}(0,1),
\end{equation}
where $\sigma_u$ is the empirical standard deviation of the field. For
complex-valued fields, the real and imaginary components are perturbed
independently using their own empirical standard deviations:
\begin{equation}
  u^{\mathrm{noisy}}
  = u
  + \frac{\rho}{100}\,\sigma_{\Re u}\,\eta_{\Re}
  + i\,\frac{\rho}{100}\,\sigma_{\Im u}\,\eta_{\Im}.
\end{equation}

For structured two-dimensional grids, the noisy field is then denoised by a
two-dimensional discrete wavelet transform using the \texttt{sym8} wavelet \cite{daubechies1992ten}.
The decomposition level is
\begin{equation}
  J = \min\{J_{\max},4\},
\end{equation}
where $J_{\max}$ is the maximum level allowed by the smaller grid dimension.
Let
\begin{equation}
  \mathcal{W}u^{\mathrm{noisy}}
  =
  \left(A_J,\{H_j,V_j,D_j\}_{j=1}^{J}\right)
\end{equation}
be the 2D wavelet coefficients, where $A_J$ is the approximation block and
$H_j,V_j,D_j$ are horizontal, vertical, and diagonal detail blocks. The noise
standard deviation is estimated from the finest diagonal detail coefficients by
the median absolute deviation estimator
\begin{equation}
  \widehat{\sigma}_n =
  \frac{\operatorname{median}(|D_1|)}{0.6745},
  \qquad
  \widehat{\sigma}_n^2 = \widehat{\sigma}_n^{\,2}.
\end{equation}
For each detail block $C \in \{H_j,V_j,D_j\}$, BayesShrink \cite{chang2000bayesshrink} estimates the
signal standard deviation as
\begin{equation}
  \widehat{\sigma}_x(C)
  =
  \sqrt{\max\!\left(\frac{1}{|C|}\sum_{m} C_m^2
  - \widehat{\sigma}_n^2,\,10^{-12}\right)}
\end{equation}
and uses the threshold
\begin{equation}
  \tau(C)
  =
  \begin{cases}
  \widehat{\sigma}_n^2 / \widehat{\sigma}_x(C),
    & \widehat{\sigma}_x(C) \geq 10^{-10},\\
  \max_m |C_m|,
    & \widehat{\sigma}_x(C) < 10^{-10}.
  \end{cases}
\end{equation}
The detail coefficients are soft-thresholded \cite{donoho1995adapting},
\begin{equation}
  \mathcal{S}_{\tau}(c)
  =
  \operatorname{sign}(c)\,\max(|c|-\tau,0),
\end{equation}
and the denoised field is reconstructed by the inverse wavelet transform:
\begin{equation}
  \begin{aligned}
  u^{\mathrm{den}}
  =\mathcal{W}^{-1}\bigl(A_J,\{&
  \mathcal{S}_{\tau(H_j)}(H_j),\\
  &\mathcal{S}_{\tau(V_j)}(V_j),
  \mathcal{S}_{\tau(D_j)}(D_j)\}_{j=1}^{J}\bigr).
  \end{aligned}
\end{equation}
For complex fields, this denoising is applied separately to $\Re(u)$ and
$\Im(u)$ and the complex field is reconstructed afterward. For unstructured
point clouds and masked three-dimensional grids, noise is added but wavelet
denoising is skipped because the wavelet basis assumes a structured rectangular
grid.

\paragraph{Agent 2: Phenomenology Extractor.}

The Phenomenology Extractor instantiates $p(\mathcal{Z}_2\mid\mathcal{Z}_1)$ using a vision language model (VLM) \cite{liu2023llava} that maps diagnostic visual evidence $\mathcal{V}\in\mathcal{Z}_1$ and metadata $m$ from Agent~1 into structured physical regime descriptors $\mathcal{Z}_2=\{c_j\}_{j=1}^{M}$, realized as $\mathcal{Z}_2 \sim q_2(\mathcal{Z}_2\mid\mathcal{Z}_1,m,\phi_1)$ for fixed prompt template $\phi_1$. This variational distribution acts as a data driven filter that concentrates probability mass on mechanistically plausible operator classes. The Extractor is prohibited from naming PDEs or generating symbolic expressions, enforced via a fixed JSON schema restricting output to structured evidence fields: variable identities, data geometry, panel level visual characteristics, transport and motion cues, structural signatures, candidate physical mechanisms (transport, dispersion, diffusion like smoothing, source sink behavior, variable coefficient effects, instability growth), and coefficient variation patterns.

\paragraph{Agent 3: Governing Law Synthesizer.}

The Governing Law Synthesizer instantiates the conditional inference term $p(\mathcal{Z}_3\mid\mathcal{Z}_2,\mathcal{D})$ as a large language model (LLM) that generates a slate of candidate governing equations. We define $\mathcal{Z}_3 = \{\mathcal{F}_k\}_{k=1}^{K},$ where each $\mathcal{F}_k$ is a symbolic hypothesis. Its variational realization is

\begin{equation}
\mathcal{F}_k \sim q_3(\mathcal{F}\mid\mathcal{Z}_2,\mathcal{D}_s,\phi_2),
\ k=1,\ldots,K,
\label{eq:proposal}
\end{equation}

where $\mathcal{D}_s\subset\mathcal{D}$ denotes a numerical subsample of the data, $\phi_2$ is the prompting template, and $\mathcal{Z}_2$ provides structured semantic conditioning that biases generation toward mechanistically plausible operator families. Unlike sparse regression pipelines that select terms from a predefined dictionary, the Synthesizer generates symbolic hypotheses \emph{de novo}. No fixed candidate library is assumed at any stage, allowing the search to consider operator compositions and variable coefficient structures proposed by the model. Each candidate is emitted in the constrained symbolic schema

\[
\begin{aligned}
\{&\texttt{EQ\_ID},\,\texttt{TARGET\_TERM},\\
  &\texttt{FEATURE\_TERMS},\,\texttt{CANONICAL\_PDE}\},
\end{aligned}
\]

where the target term carries an implicit unit coefficient and feature terms encode differential operators, polynomial nonlinearities (e.g., $u^2$, $u\,u_x$), and, when supported by upstream evidence, coordinate dependent coefficients such as $x\cdot u_x$. Coefficient estimation is explicitly deferred to the Equation Arbiter.

At each iteration, the Synthesizer emits exactly ten structurally distinct candidates. To reduce premature mode collapse, no more than two may refine the previously best hypothesis, while at least four must originate from materially different structural families; this diversity constraint retains unresolved alternatives for downstream arbitration. The resulting $\mathcal{Z}_3=\{\mathcal{F}_k\}_{k=1}^{K}$ is a mechanistically conditioned, library free hypothesis frontier structured for numerical falsification.

\paragraph{Agent 4: Equation Arbiter.}

The Equation Arbiter instantiates the terminal validation stage of Eq.~\eqref{eq:factorization}. For each candidate $\mathcal{F}_k$ and prompt template $\phi_3$, Agent 4 (LLM) generates validation code $\mathcal{C}_k \sim q_4(\mathcal{C}\mid\mathcal{F}_k, \phi_3),$ which estimates coefficients and residual loss,

\begin{equation}
\boldsymbol{\theta}_k^{\ast},\;\mathcal{L}_k
=
\mathrm{Exec}(\mathcal{C}_k,\mathcal{D}),
\end{equation}

inducing likelihood $p(\mathcal{D}\mid\mathcal{F}_k,\boldsymbol{\theta}_k^{\ast}) \propto \exp(-\mathcal{L}_k).$ Candidates are evaluated exactly as proposed using $L_2$-normalized ordinary least squares on a deterministic 80/20 train test split, with weak form derivative estimation throughout for numerical stability (refer to Appendix). To balance fidelity and parsimony, candidate $j$ is scored by

\begin{equation}
R_j
=
R^2_{\mathrm{test},j}
\max\!\left(0,\;1-0.3\log_{10}N_j\right),
\label{eq:scoring}
\end{equation}

and ranked accordingly. The top candidate is assigned confidence

\begin{equation}
C
=
\max\!\left(0,\;1-\sqrt{\max(0,1-R^2_{\mathrm{test}})}\right)\cdot100.
\end{equation}

Acceptance requires $C\geq80$; otherwise, the slate is rejected and terminal reward is propagated backward for regeneration. This falsification step provides a practical safeguard against accepting weakly supported candidates.

\subsubsection{Weak-Form Formulation}
\label{sec:appendix_weak_form_validation}
Agent~4 validates each candidate PDE by converting its target term and feature
terms into weak-form support integrals. For a structured grid with independent
coordinates $r=(r_1,\ldots,r_d)$, each support is centered at $r_c$ and uses
normalized local coordinates
\begin{equation}
  s_i = \frac{r_i-r_{c,i}}{L_i}, \qquad s_i \in [-1,1],
  \qquad L_i = H_i\,\Delta r_i,
\end{equation}
where $\Delta r_i$ is the grid spacing and $H_i$ is the chosen half-width in
grid cells. The compact one-dimensional test function is
\begin{equation}
  \phi(s) = (1-s^2)^{2p}, \qquad p=4,
\end{equation}
and the multidimensional test function is separable:
\begin{equation}
  \Phi(s_1,\ldots,s_d) = \prod_{i=1}^{d}\phi(s_i).
\end{equation}
Derivatives of $\phi$ are computed analytically from the polynomial
coefficients, not by finite differences.

For a derivative term $D^\alpha u$ with multi-index
$\alpha=(\alpha_1,\ldots,\alpha_d)$ and
$|\alpha|=\sum_i \alpha_i$, integration by parts gives the weak contribution
used in the regression row:
\begin{equation}
  \begin{aligned}
  \mathcal{I}_{\alpha}[u]
  ={}&(-1)^{|\alpha|}
  \left(\prod_{i=1}^{d} L_i^{\,1-\alpha_i}\right)
  \int_{[-1,1]^d} u(r_c + L\odot s)\\
  &{}\times
  \prod_{i=1}^{d}\phi^{(\alpha_i)}(s_i)\,ds.
  \end{aligned}
  \label{eq:weak_form_contribution}
\end{equation}
The factor $(-1)^{|\alpha|}$ is the integration-by-parts sign, and
$L_i^{1-\alpha_i}$ combines the physical-coordinate Jacobian with the
derivative scaling. The integral in
Eq.~\eqref{eq:weak_form_contribution} is evaluated by trapezoidal quadrature
over the normalized support grid. If the data contain masked or invalid values
(\texttt{NaN}), a support is used only when the integration stencil remains on
valid finite data; the validator is instructed not to integrate across masked
regions.

For candidate $j$, let $T_j$ be its target term and
$F_{j,1},\ldots,F_{j,m_j}$ be its feature terms. Each valid compact support
produces one row of the regression system
\begin{equation}
  \begin{aligned}
  y_j[q] &= \mathcal{I}_{T_j}^{(q)}[u],\\
  \Theta_j[q,k] &= \mathcal{I}_{F_{j,k}}^{(q)}[u],
  \qquad k=1,\ldots,m_j.
  \end{aligned}
\end{equation}
so that the fitted equation is obtained from
\begin{equation}
  y_j \approx \Theta_j \xi_j.
\end{equation}
The same valid support-row intersection and the same deterministic 80/20
train-test split are reused for all candidates. Columns of
$\Theta_{\mathrm{train}}$ are normalized by their training-set $\ell_2$ norms;
least squares is solved on the normalized system, and the coefficients are then
rescaled back to physical units. If $u$, $y_j$, or $\Theta_j$ is complex-valued,
the same construction is carried out in complex arithmetic. Candidate ranking
is based on held-out residual quality with a sparsity penalty, so the accepted
equation must both predict unseen weak-form rows and remain parsimonious.

For unstructured point-cloud data, Agent~4 does not use the compact weak-form
grid integral. Instead, it uses local weighted polynomial least squares over
$k$ nearest neighbors. If the polynomial degree is $p$ in dimension $d$, the
local monomial count is
\begin{equation}
  n_{\mathrm{mono}} = \binom{p+d}{d},
\end{equation}
and the neighborhood size is chosen with
$k \geq 2 n_{\mathrm{mono}}$. Requested derivatives are read analytically from
the fitted local polynomial coefficients as $\alpha!$ times the coefficient of
the corresponding centered monomial.

\subsection{Autonomous Hypothesis Validation Loop}

The four agents operate within a closed loop refinement process driven by validation feedback from the Equation Arbiter. Following arbitration, the system selects one of three deterministic actions,

\[
a \in \{\mathrm{accept},\,\mathrm{refine},\,\mathrm{restart}\},
\]

according to the confidence score returned by the Equation Arbiter. When validation fails, control first enters an inner refinement loop that routes feedback directly to the Governing Law Synthesizer. The rejected slate, associated validation scores, and validator diagnostics are appended to the interaction history, which condition the next proposal round as,

\begin{equation}
\mathcal{F}_k^{(t+1)}
\sim
q_3\!\left(
\mathcal{F}
\mid
\mathcal{Z}_2,
\mathcal{D}_s,
\phi_2,
\mathcal{Z}_4^{(t)}
\right),\ k=1,\ldots,K.
\label{eq:refine}
\end{equation}

This feedback can steer generation away from previously rejected structural families while preserving diversity constraints. Validator runtime failures trigger code regeneration without altering the candidate slate. If refinement exhausts its budget, control enters an outer restart loop that re-invokes the Phenomenology Extractor to refresh the semantic prior, followed by a new synthesis round. The highest scoring candidate discovered so far is retained across restarts. Together, these nested loops implement an iterative falsification driven refinement process that encourages later proposals to be more mechanically plausible and numerically consistent. Unlike a static single-pass pipeline, symbolic generation remains conditioned on numerical validation feedback until acceptance or budget exhaustion.
\section{Results and Discussion}
\label{sec:results}

\begin{figure}[htbp] 
\centering 
\includegraphics[width=\textwidth]{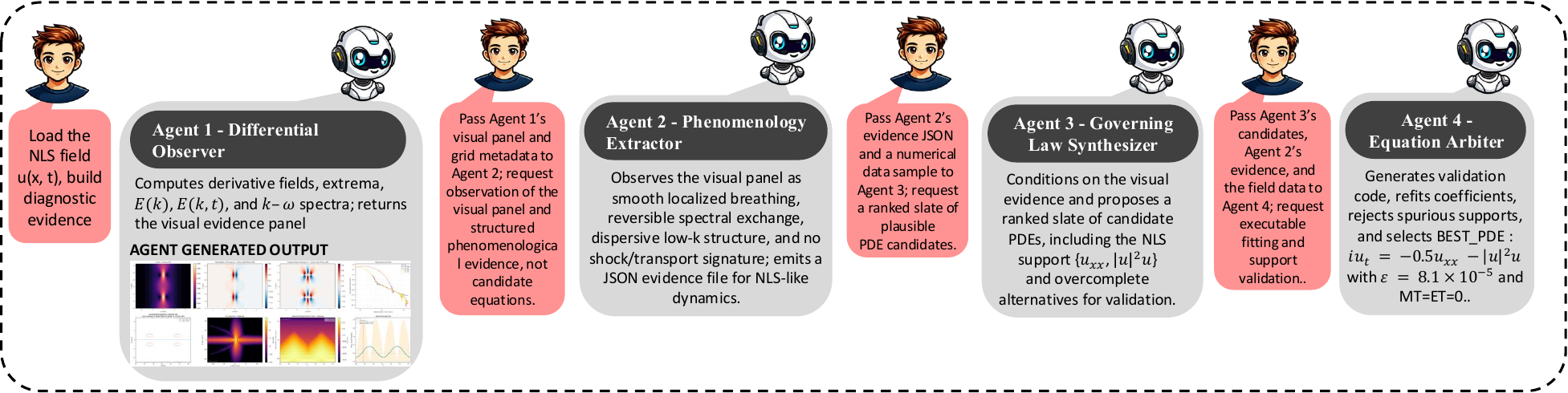} \caption{End to end flow of the multimodal agentic framework MAGE on the NLS benchmark. The user/system controller routes artifacts between specialized agents: diagnostic visual evidence from Agent~1, a JSON evidence file from Agent~2, candidate equations from Agent~3, and executable validation from Agent~4. Detailed prompts and intermediate outputs are provided; refer to Appendix.} \label{fig:mage_in_action_compact} 
\end{figure}

Figure~\ref{fig:mage_in_action_compact} illustrates this end to end observation hypothesis verification loop, where diagnostic features are extracted, physical regimes inferred, candidate equations synthesized, and numerical validation determines acceptance or rejection. The experiments below examine this workflow on canonical PDE benchmarks, two complex geometries, additive-noise perturbations, and a laboratory sensor record, together with targeted component ablations.
\FloatBarrier

\subsection{Canonical PDE Benchmark Suite}
\label{sec:time_dependent}

Table~\ref{tab:aggregate} combines aggregate performance with per-equation results across the evaluated canonical PDEs~\cite{rudy2017data,xu2025eqgpt}. It reports two complementary dimensions: (i) symbolic structure recovery, measured through exact recovery, structural failure, missing terms (MT), and extra terms (ET), and (ii) coefficient estimation accuracy, measured through normalized RMS coefficient error. All library based baselines are evaluated using a unified candidate library containing all ground truth terms, ensuring comparison within a shared hypothesis space.

Classical sparse regression methods (PF~\cite{rudy2017data}, SGA~\cite{chen2022symbolic}, WS~\cite{messenger2021weak}, WI~\cite{tang2023weakident}) achieve 14.3--42.9\% exact recovery in this comparison, with coefficient instability and structural failures in several nonlinear and dispersive cases; the tested physics informed and hybrid approaches (PSR~\cite{chen2021physics}, PR~\cite{stephany2022pderead}) recover no exact structure here. Because the compared library based methods have access to all ground truth terms, these failures suggest that composition and estimation within the shared symbolic space remain difficult on this suite.

\begin{table}[htbp]
\centering
{%
\footnotesize
\renewcommand{\arraystretch}{1.08}
\setlength{\tabcolsep}{0.18mm}
\newcommand{\benchmetric}[1]{\multicolumn{1}{>{\centering\arraybackslash}m{0.95cm}}{#1}}
\newcommand{\bencheq}[2]{\multirow{3}{3.7cm}[-0.30em]{\centering\makecell[c]{\textbf{#1}\\[-0.1em]\scriptsize $#2$}}}
\newcommand{\benchsp}[1]{\cellcolor{MageSpurious}#1\textsuperscript{S}}
\newcommand{\benchfail}[1]{\cellcolor{MageBad}#1}
\begin{tabular}{@{}
>{\raggedright\arraybackslash}m{3.7cm}
>{\centering\arraybackslash}m{0.95cm}
*{8}{>{\centering\arraybackslash}m{1.15cm}}
>{\columncolor{MageGood}\centering\arraybackslash}m{1.15cm}
@{}}
\toprule
\mageTableHead
\multicolumn{2}{l}{\textbf{Aggregate summary}}
& \textbf{PF} & \textbf{SGA} & \textbf{WS} & \textbf{WI}
& \textbf{PSR} & \textbf{PR} & \textbf{L4E} & \textbf{EG} & \textbf{MAGE} \\
\midrule
\multicolumn{2}{l}{\textbf{Exact (\%)}}
& 42.9 & 14.3 & 28.6 & 42.9 & 0.0 & 0.0 & 85.7 & 71.4 & \textbf{100.0} \\
\multicolumn{2}{l}{\textbf{Fail (\%)}}
& 57.1 & 85.7 & 71.4 & 57.1 & 100.0 & 100.0 & 14.3 & 28.6 & \textbf{0.0} \\
\multicolumn{2}{l}{\textbf{$\overline{\varepsilon}$}}
& 5.04e-2 & 4.70e-2 & 2.18e-2 & 8.90e-4 & 2.49e2 & 4.49e4 & 3.13e-2 & 6.93e-2 & \textbf{7.96e-7} \\
\multicolumn{2}{l}{\textbf{Characteristic}}
& Spur. & High-$\varepsilon$ & Unstable & Sensitive & Severe & Failed & Strong & Moderate & \textbf{Best tested} \\
\midrule
\mageTableHead
\multicolumn{2}{l}{\textbf{Per-equation results}}
& \multicolumn{9}{c}{\textbf{Method}} \\
\cmidrule(lr){3-11}
\mageTableHead
\multicolumn{1}{c}{\textbf{Equation}} & \textbf{Metric}
& \textbf{PF} & \textbf{SGA} & \textbf{WS} & \textbf{WI}
& \textbf{PSR} & \textbf{PR} & \textbf{L4E} & \textbf{EG} & \textbf{MAGE} \\
\midrule

\bencheq{NLS}{iu_t={-}0.5u_{xx}{-}|u|^2u}
& \benchmetric{$\varepsilon$}
& \benchsp{1.33e-1} & \benchsp{0.11247} & 1.79e-4 & \benchfail{1.12}
& \benchsp{0.106} & \benchsp{0.139} & \benchfail{0.788} & \benchfail{1.205} & \textbf{8.10e-5} \\
& \benchmetric{MT} & \textbf{0} & \textbf{0} & \textbf{0} & 2 & \textbf{0} & \textbf{0} & 1 & 3 & \textbf{0} \\
& \benchmetric{ET} & 11 & 5 & \textbf{0} & 1 & 2 & 2 & 1 & 3 & \textbf{0} \\
\midrule

\bencheq{PDE-Divide}{u_t=0.25u_{xx}-u_x/x}
& \benchmetric{$\varepsilon$}
& \benchsp{2.48e-1} & \benchsp{9.22e-2} & \benchfail{1} & \benchfail{2.303}
& \benchfail{1.24e4} & \benchfail{9.09e5} & 9.31e-2 & 4.29e-2 & \textbf{2.36e-4} \\
& \benchmetric{MT} & \textbf{0} & \textbf{0} & 2 & 2 & 1 & 2 & \textbf{0} & \textbf{0} & \textbf{0} \\
& \benchmetric{ET} & 17 & 21 & \textbf{0} & 1 & 20 & 4 & \textbf{0} & \textbf{0} & \textbf{0} \\
\midrule

\bencheq{KdV}{u_t={-}6uu_x-u_{xxx}}
& \benchmetric{$\varepsilon$}
& 9.65e-4 & \benchsp{1.94} & \benchfail{1.76} & 1.49e-3
& \benchfail{6.76e4} & \benchfail{5.51e5} & 2.94e-3 & \benchfail{1.75} & \textbf{4.93e-6} \\
& \benchmetric{MT} & \textbf{0} & \textbf{0} & 1 & \textbf{0} & \textbf{0} & 2 & \textbf{0} & 2 & \textbf{0} \\
& \benchmetric{ET} & \textbf{0} & 3 & 7 & \textbf{0} & 25 & 12 & \textbf{0} & 3 & \textbf{0} \\
\midrule

\bencheq{KS}{u_t={-}uu_x-u_{xx}-u_{xxxx}}
& \benchmetric{$\varepsilon$}
& 6.66e-3 & \benchsp{4.56e-3} & \benchsp{1.24e-3} & \benchfail{0.600314}
& \benchfail{1.49e1} & \benchfail{4.90e6} & 2.61e-1 & \benchfail{9.94e-1} & \textbf{1.73e-6} \\
& \benchmetric{MT} & \textbf{0} & \textbf{0} & \textbf{0} & 1 & 3 & 3 & \textbf{0} & 2 & \textbf{0} \\
& \benchmetric{ET} & \textbf{0} & 1 & 1 & \textbf{0} & 20 & 10 & \textbf{0} & 3 & \textbf{0} \\
\midrule

\bencheq{Allen-Cahn}{u_t=0.003u_{xx}+u-u^3}
& \benchmetric{$\varepsilon$}
& \benchsp{9.37e-1} & \benchsp{1.36e1} & \benchfail{6.23e-1} & \benchsp{8.58e-1}
& \benchfail{6.59e2} & \benchfail{2.23e6} & 1.02e-3 & 2.48e-3 & \textbf{7.07e-7} \\
& \benchmetric{MT} & \textbf{0} & \textbf{0} & 1 & \textbf{0} & 1 & 1 & \textbf{0} & \textbf{0} & \textbf{0} \\
& \benchmetric{ET} & 1 & 2 & \textbf{0} & 2 & 5 & 11 & \textbf{0} & \textbf{0} & \textbf{0} \\
\midrule

\bencheq{Burgers}{u_t={-}uu_x+0.1u_{xx}}
& \benchmetric{$\varepsilon$}
& \benchfail{1.52e-1} & 1.01e-3 & 1.06e-3 & 1.06e-3
& \benchfail{1.50e3} & \benchfail{1.42e6} & 2.59e-3 & 2.90e-3 & \textbf{8.02e-6} \\
& \benchmetric{MT} & 1 & \textbf{0} & \textbf{0} & \textbf{0} & 2 & 2 & \textbf{0} & \textbf{0} & \textbf{0} \\
& \benchmetric{ET} & \textbf{0} & \textbf{0} & \textbf{0} & \textbf{0} & 16 & 20 & \textbf{0} & \textbf{0} & \textbf{0} \\
\midrule

\bencheq{Conv.-Diffusion}{u_t=0.25u_{xx}-u_x}
& \benchmetric{$\varepsilon$}
& 5.22e-4 & \benchsp{1.35e-2} & \benchfail{1} & \textbf{0}
& \benchfail{3.54e2} & \benchfail{1.67} & 1.82e-2 & 5.76e-2 & 1.75e-4 \\
& \benchmetric{MT} & \textbf{0} & \textbf{0} & 2 & \textbf{0} & 1 & 2 & \textbf{0} & \textbf{0} & \textbf{0} \\
& \benchmetric{ET} & \textbf{0} & 2 & \textbf{0} & \textbf{0} & 18 & 1 & \textbf{0} & \textbf{0} & \textbf{0} \\
\midrule

\bencheq{Wave}{u_{tt}=u_{xx}}
& \benchmetric{$\varepsilon$}
& \benchsp{2.6385} & 1.40e-3 & 1.95e-4 & 1.88e-4
& \benchsp{31.96} & \benchfail{9.09e6} & \benchsp{0.339} & 1.42e-2 & \textbf{0} \\
& \benchmetric{MT} & \textbf{0} & \textbf{0} & \textbf{0} & \textbf{0} & \textbf{0} & 1 & \textbf{0} & \textbf{0} & \textbf{0} \\
& \benchmetric{ET} & 3 & \textbf{0} & \textbf{0} & \textbf{0} & 22 & 14 & 1 & \textbf{0} & \textbf{0} \\
\bottomrule
\end{tabular}
\par\smallskip
\begin{minipage}{0.94\textwidth}
\scriptsize
\textbf{Color key:}\quad
\colorbox{MageSpurious}{\strut\hspace{2.1em}}\;spurious discovered structure (MT=0, ET$>$0)\qquad
\colorbox{MageBad}{\strut\hspace{2.1em}}\;structurally incorrect recovery (MT$>$0)\qquad
\colorbox{MageGood}{\strut\hspace{2.1em}}\;MAGE results
\end{minipage}
}
\caption{Aggregate and per-equation quantitative comparison on the canonical PDE benchmarks. $\overline{\varepsilon}$ is the geometric mean of the eight displayed per-equation normalized RMS coefficient errors $\varepsilon$, including errors for structurally incorrect recoveries; exact zeros are replaced by $10^{-15}$ for logarithmic averaging. MT and ET denote missing and extra term counts. Lower is better for all three error metrics, and bold indicates the best value in each row. PF: PDE-FIND; WS: W-SINDy; WI: Weak-Ident; PSR: PINN-SR; PR: PDE-READ; L4E: LLM4ED; EG: EqGPT.}
\label{tab:aggregate}
\end{table}
\FloatBarrier

LLM based methods (L4E~\cite{du2024llm4ed}, EG~\cite{xu2025eqgpt}) improve structural recovery in this comparison, reaching 85.7\% and 71.4\% exact recovery, respectively, though residual coefficient errors remain in stiff and coupled cases. MAGE obtains 100.0\% exact recovery with zero structural failures and approximately 3 orders of magnitude lower geometric-mean coefficient error on the evaluated suite. Beyond exact structural accuracy, MAGE delivers the strongest coefficient fidelity on seven of eight equations, often by substantial margins. The gains are particularly pronounced on structurally challenging systems such as KS, Allen--Cahn, and PDE-Divide, where MAGE reduces coefficient error by multiple orders of magnitude relative to the strongest competing method. The baseline failure modes are less evident in these runs, which is consistent with, but does not by itself isolate, the contribution of closed loop falsification driven refinement.

\subsection{Experimental Sensor Data}
\label{sec:experimental_sensor_data}

The experimental case in the main paper uses the Silverbox dataset, a
measured input--output record from an analog electronic circuit designed to
approximate a nonlinear mechanical resonator. The acquisition and
signal-generation cards were clock-synchronized, and both channels were sampled
at $610.35$\,Hz~\cite{wigren2013three}. In the dataset's idealized physical
description, the response $y$ satisfies
\begin{equation}
  m y_{tt} + d y_t + k(y)y = f,
  \qquad k(y)=a+b y^2,
  \label{eq:silverbox_idealized}
\end{equation}
so the position-dependent stiffness produces both $y$ and $y^3$ restoring
terms. The source also notes that the electronic circuit is close to, but not
exactly equal to, this idealization. The dataset therefore tests recovery from
experimental measurements with model mismatch rather than from data generated
exactly by the target equation.

The official archive identifies \texttt{V1} as the applied input and
\texttt{V2} as the measured output, and includes an additional
Schroeder-phase multisine record~\cite{nonlinearbenchmark2026silverbox}. We use
that release's \texttt{Schroeder80mV} record and map its channels to the main
paper's notation as $f(t):=\texttt{V1}(t)$ and
$u(t):=\texttt{V2}(t)$. Table~\ref{tab:silverbox_split} records the exact
project-specific temporal split.

\begin{center}
\centering
\setlength{\tabcolsep}{6pt}
{\small
\mageTableRows
\renewcommand{\arraystretch}{1.12}
\begin{tabular}{@{}lrrl@{}}
\toprule
\mageTableHead
\textbf{Partition} & \textbf{Samples} & \textbf{Time (s)} & \textbf{Role} \\
\midrule
Estimation & $10{,}400$ & $0.139$--$17.177$ & Fit coefficients \\
Held-out & $120{,}587$ & $17.179$--$214.747$ & Rank and evaluate \\
\bottomrule
\end{tabular}
}
\captionsetup{hypcap=false}
\captionof{table}{Temporal partition of the \texttt{Schroeder80mV} experimental record.}
\label{tab:silverbox_split}
\end{center}

Second-order finite-difference derivatives are computed separately within the
two partitions so that no derivative stencil crosses the temporal boundary.
Candidate coefficients are estimated only on the estimation partition using
column-normalized ordinary least squares, and fitted equations are evaluated on
the held-out partition without coefficient refitting. Because the same
held-out partition is also used to rank candidate structures, it is a holdout
for coefficient estimation but not a fully untouched final test record. This
distinction scopes the experimental result and motivates future validation on
an additional independent record or operating regime.

Ranking by held-out error selects the cubic model
\begin{equation}
\begin{split}
u_{tt}={}&-24.2035u_t-152423u\\
          &-401629u^3+97359.6f,
\end{split}
\label{eq:experimental_cubic}
\end{equation}
with $R^2=0.98538$, NRMSE $=0.12090$, and confidence $=87.91\%$. The closest linear alternative, $u_{tt}=-24.1687u_t-157412u+97273.5f$, yields NRMSE $=0.12684$ and confidence $=87.32\%$. Although the sparsity-weighted reward favors the simpler linear form, the held-out error ranking retains the cubic term. The small metric gap cautions against a broad robustness claim, but recovery of the expected nonlinearity and coefficient signs provides initial evidence that the workflow can remain useful on noisy sensor observations outside the canonical PDE suite.

\subsection{Complex Geometric Domains}
\label{sec:time_independent}

\begin{figure}[htbp]
\centering
\begin{tikzpicture}
\node[draw, densely dotted, thick, rounded corners=1ex, inner sep=0.5ex] (box) {%
\begin{minipage}{0.98\textwidth}
\vspace{0.4ex}

\begin{minipage}[t]{0.48\textwidth}
\centering
\small\textbf{2D plate}\par\vspace{0.5ex}

\begin{minipage}[c]{0.48\textwidth}
\centering
\setlength{\fboxsep}{2pt}
\fcolorbox{blue!35}{blue!4}{%
\begin{minipage}{0.95\linewidth}
\centering\scriptsize
True: $u_{xx}+u_{yy}=0$\\[0.2ex]
MAGE:\\[-0.1ex]
$u_{xx}+0.990065u_{yy}=0$
\end{minipage}}
\end{minipage}
\hfill
\begin{minipage}[c]{0.48\textwidth}
\centering
\includegraphics[width=\linewidth]{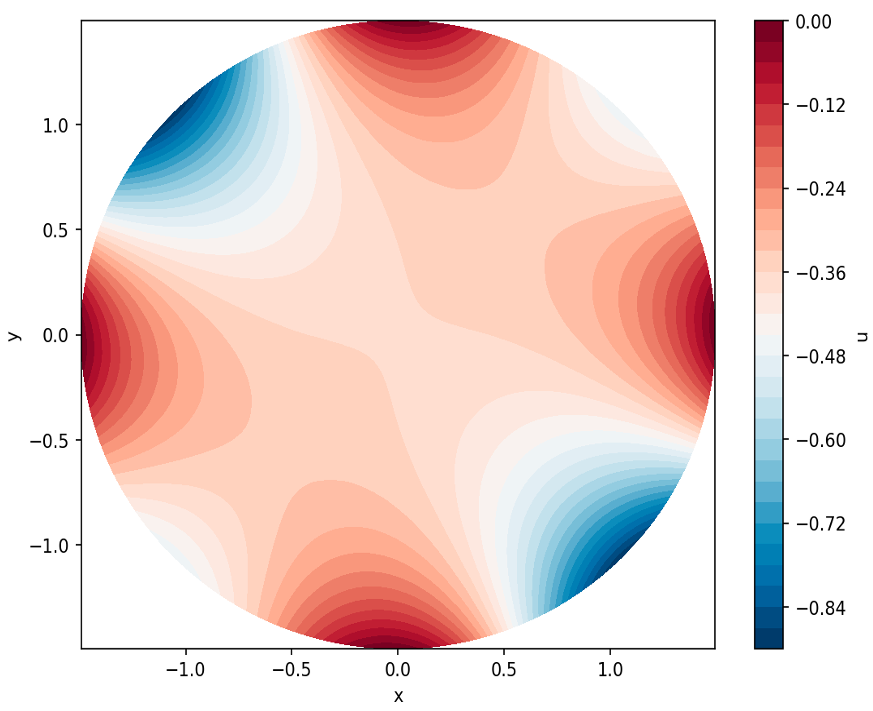}
\end{minipage}
\end{minipage}
\hfill
\begin{minipage}[t]{0.48\textwidth}
\centering
\small\textbf{3D space shuttle}\par\vspace{0.5ex}

\begin{minipage}[c]{0.42\textwidth}
\centering
\setlength{\fboxsep}{2pt}
\raisebox{-2.0ex}[0pt][0pt]{\fcolorbox{blue!35}{blue!4}{%
\begin{minipage}{0.95\linewidth}
\centering\scriptsize
True:\\[-0.1ex]
$u_{xx}+u_{yy}+u_{zz}=0$\\[0.2ex]
MAGE:\\[-0.1ex]
$u_{xx}+1.039452u_{yy}+$\\[-0.1ex]
$0.99838459u_{zz}=0$
\end{minipage}}}
\end{minipage}
\hfill
\begin{minipage}[c]{0.56\textwidth}
\centering
\includegraphics[width=\linewidth,trim=65 45 60 70,clip]{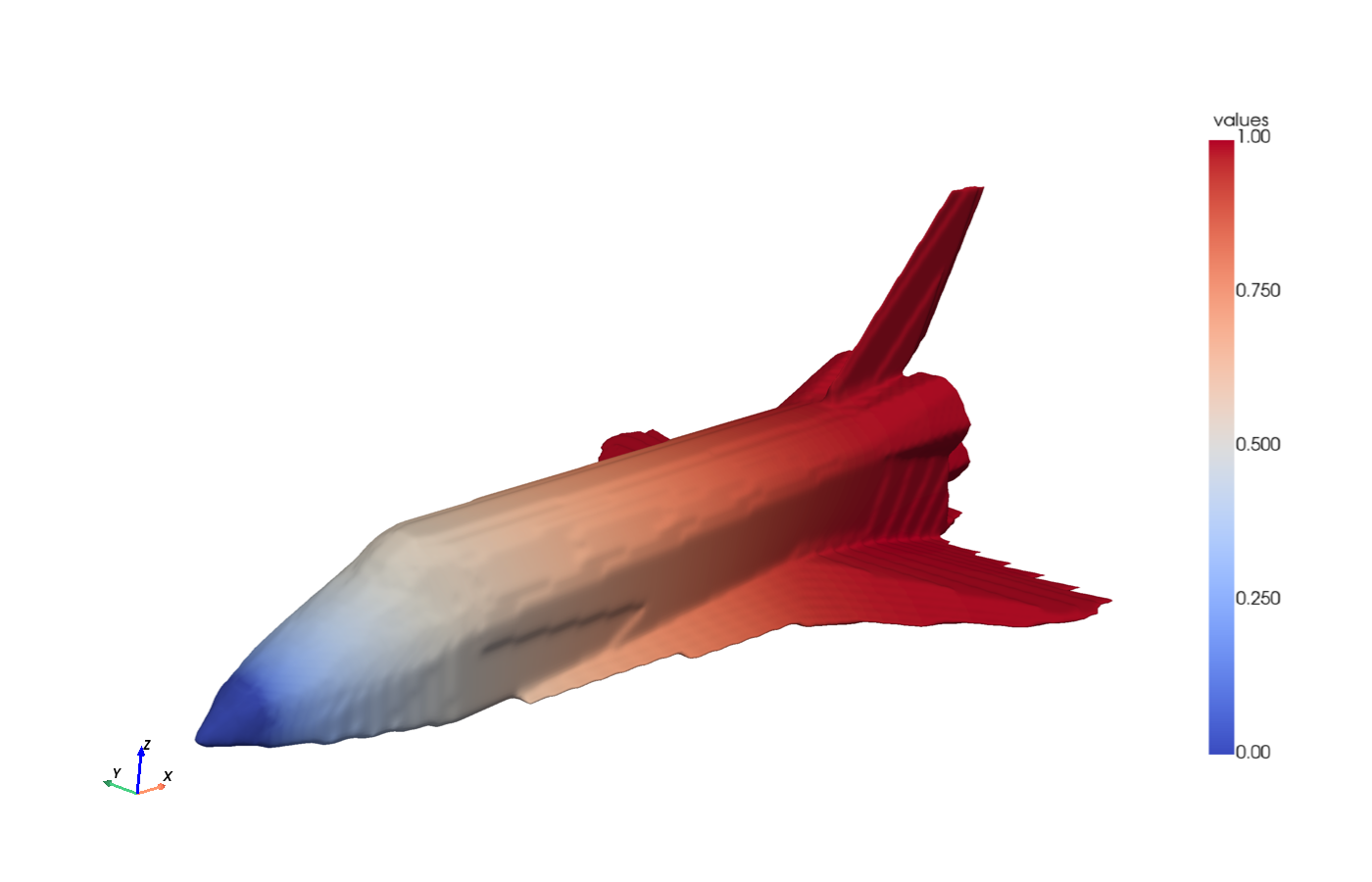}
\end{minipage}
\end{minipage}

\vspace{0.3ex}
\end{minipage}
};
\draw[densely dotted, thick] (box.north) -- (box.south);
\end{tikzpicture}
\caption{PDE discovery in complex geometries.}
\label{fig:time_independent_laplacian}
\end{figure}

Several compared sparse regression and neural surrogate methods (PDE-FIND~\cite{rudy2017data}, SGA~\cite{chen2022symbolic}, W-SINDy~\cite{messenger2021weak}, Weak-Ident~\cite{tang2023weakident}, PINN-SR~\cite{chen2021physics}, PDE-READ~\cite{stephany2022pderead}) are commonly formulated as $u_t=\Theta\hat{\boldsymbol{\xi}}$; when $u_t\equiv0$, that formulation has a degenerate temporal target and requires reformulation for steady-state discovery. MAGE instead operates directly on spatial contour observations. On a 2D plate and a 3D space shuttle~\cite{xu2025eqgpt} (Fig.~\ref{fig:time_independent_laplacian}), it recovers the Laplacian support with zero missing or spurious terms and low single digit coefficient error. These two cases suggest that the framework can identify steady elliptic structure from spatial observations, but they do not cover the broader range of elliptic operators or geometries.

\subsection{Noise Robustness}
\label{sec:noise_robustness}
Table~\ref{tab:noise_robustness} evaluates MAGE under increasing additive Gaussian noise, jointly tracking coefficient stability and structural recovery. Coefficient error $\varepsilon$ generally increases with noise and remains comparatively small through 5--10\% noise for several smooth transport/diffusion equations (KdV, Burgers, Allen-Cahn), while PDE-Divide and Conv.-Diff. degrade earlier, consistent with greater derivative-estimation sensitivity. Structural error remains zero in most tested cases until higher noise levels. This separation suggests that the validation loop can preserve structure after coefficient estimates begin to degrade, although the results are specific to the tested equations, noise model, and derivative estimator.

\begin{table}[t]
\centering
{%
\renewcommand{\arraystretch}{1.12}
\setlength{\tabcolsep}{1mm}
\newcommand{\noiseExcellent}[1]{\cellcolor{MageNoiseExcellent}#1}
\newcommand{\noiseGood}[1]{\cellcolor{MageNoiseGood}#1}
\newcommand{\noiseModerate}[1]{\cellcolor{MageNoiseModerate}#1}
\newcommand{\noiseHigh}[1]{\cellcolor{MageNoiseHigh}#1}
\newcommand{\noiseCritical}[1]{\cellcolor{MageNoiseCritical}#1}
\newcommand{\noiseClean}[1]{\cellcolor{MageNoiseClean}#1}
\newcommand{\noiseSwatch}[1]{{\setlength{\fboxsep}{0pt}\colorbox{#1}{\rule{0pt}{1.3ex}\rule{1.6ex}{0pt}}}}
\begin{tabular}{l *{6}{>{\centering\arraybackslash}m{1.08cm}} *{6}{>{\centering\arraybackslash}m{0.78cm}}}
\toprule
\mageTableHead
& \multicolumn{6}{c}{\textbf{Coefficient error}\,\,$\boldsymbol{\varepsilon}$}
& \multicolumn{6}{c}{\textbf{Structural error}\,\,(\textbf{MT}$+$\textbf{ET})} \\
\cmidrule(lr){2-7}\cmidrule(lr){8-13}
\mageTableHead
\textbf{Equation}
& \textbf{0\%} & \textbf{1\%} & \textbf{2\%} & \textbf{5\%} & \textbf{10\%} & \textbf{15\%}
& \textbf{0\%} & \textbf{1\%} & \textbf{2\%} & \textbf{5\%} & \textbf{10\%} & \textbf{15\%} \\
\midrule
NLS
  & \noiseExcellent{8.10e-5} & \noiseGood{1.2e-4} & \noiseGood{2.4e-4} & \noiseModerate{1.21e-3} & \noiseModerate{3.63e-3} & \noiseCritical{6.34e-1}
  & \noiseClean{0} & \noiseClean{0} & \noiseClean{0} & \noiseClean{0} & \noiseClean{0} & \noiseClean{0} \\
PDE-Divide
  & \noiseGood{2.36e-4} & \noiseHigh{1.71e-2} & \noiseCritical{2.96e0} & \noiseCritical{3.10e0} & \noiseCritical{3.09e0} & \noiseCritical{3.34e0}
  & \noiseClean{0} & \noiseClean{0} & \noiseCritical{3} & \noiseCritical{3} & \noiseCritical{5} & \noiseCritical{5} \\
KdV
  & \noiseExcellent{4.93e-6} & \noiseGood{1.0e-4} & \noiseGood{7.5e-4} & \noiseModerate{7.95e-3} & \noiseHigh{3.28e-2} & \noiseCritical{1.44e-1}
  & \noiseClean{0} & \noiseClean{0} & \noiseClean{0} & \noiseClean{0} & \noiseClean{0} & \noiseClean{0} \\
KS
  & \noiseExcellent{1.73e-6} & \noiseGood{9.1e-4} & \noiseModerate{1.75e-3} & \noiseModerate{5.12e-3} & \noiseModerate{1.79e-3} & \noiseCritical{2.16e-1}
  & \noiseClean{0} & \noiseClean{0} & \noiseClean{0} & \noiseClean{0} & \noiseCritical{1} & \noiseClean{0} \\
Burgers
  & \noiseExcellent{8.0e-6} & \noiseModerate{1.46e-3} & \noiseModerate{3.31e-3} & \noiseModerate{1.15e-3} & \noiseHigh{2.88e-2} & \noiseHigh{1.35e-2}
  & \noiseClean{0} & \noiseClean{0} & \noiseClean{0} & \noiseCritical{1} & \noiseClean{0} & \noiseCritical{1} \\
Allen-Cahn
  & \noiseExcellent{7.07e-7} & \noiseModerate{1.11e-3} & \noiseModerate{2.20e-3} & \noiseModerate{4.01e-3} & \noiseModerate{5.84e-3} & \noiseHigh{1.24e-2}
  & \noiseClean{0} & \noiseClean{0} & \noiseClean{0} & \noiseClean{0} & \noiseClean{0} & \noiseClean{0} \\
Wave
  & \noiseExcellent{0} & \noiseExcellent{0} & \noiseExcellent{3e-6} & \noiseHigh{1.01e-2} & \noiseHigh{2.07e-2} & \noiseHigh{2.69e-2}
  & \noiseClean{0} & \noiseClean{0} & \noiseClean{0} & \noiseClean{0} & \noiseClean{0} & \noiseCritical{2} \\
Conv.-Diff.
  & \noiseGood{1.75e-4} & \noiseModerate{3.43e-3} & \noiseModerate{4.56e-3} & \noiseHigh{5.94e-2} & \noiseHigh{2.63e-2} & \noiseCritical{3.43e-1}
  & \noiseClean{0} & \noiseClean{0} & \noiseClean{0} & \noiseCritical{1} & \noiseCritical{1} & \noiseCritical{3} \\
\bottomrule
\end{tabular}
\par\smallskip
\begin{minipage}{\textwidth}
\scriptsize
\textbf{Coefficient-error scale:}\quad
\noiseSwatch{MageNoiseExcellent}\;$<10^{-4}$\quad
\noiseSwatch{MageNoiseGood}\;$10^{-4}$--$10^{-3}$\quad
\noiseSwatch{MageNoiseModerate}\;$10^{-3}$--$10^{-2}$\quad
\noiseSwatch{MageNoiseHigh}\;$10^{-2}$--$10^{-1}$\quad
\noiseSwatch{MageNoiseCritical}\;$\geq10^{-1}$
\qquad\textbar\qquad
\textbf{Structural error:}\quad
\noiseSwatch{MageNoiseClean}\;$0$\quad
\noiseSwatch{MageNoiseCritical}\;$\geq1$
\end{minipage}
}
\caption{Noise robustness under additive Gaussian perturbations. Each cell reports the measured value, while the background encodes its severity tier. The left block gives coefficient error $\varepsilon$ on a logarithmic scale; the right block gives structural error (MT$+$ET), which remains zero through 1\% noise for all equations, with the first failure appearing for PDE-Divide at 2\%.}
\label{tab:noise_robustness}
\end{table}

\subsection{LLM/VLM Sensitivity Study}
We evaluate sensitivity to foundation model choice by varying the VLM (Phenomenology Extractor) and LLM (Governing Law Synthesizer) on the NLS case. Table~\ref{tab:llm_vlm_sensitivity} summarizes the results, with full cue level rankings in the Appendix. Multiple tested backbones recover the target structure: KIMI-K2.5 \cite{kimi2025k2} yields the strongest VLM score (4.8/5), while GLM-5.1 \cite{glmteam2024chatglm} gives the most reliable rank-1 recovery and highest true-term recall. This reduces concern that the reported NLS result depends on one backbone, but it does not establish model independence across other PDE families.

\begin{table}[htbp]
\centering
{%
\small
\setlength{\tabcolsep}{7pt}
\renewcommand{\arraystretch}{1.42}
\begin{tabularx}{\textwidth}{@{}
  >{\raggedright\arraybackslash\bfseries\color{MageNavy}}p{0.10\textwidth}
  >{\raggedright\arraybackslash\columncolor{MageGood}}X
  >{\raggedright\arraybackslash\columncolor{MageAccent}}X
  >{\raggedright\arraybackslash\columncolor{MageWarn}}X@{}}
\toprule
\mageTableHead
\textbf{Axis} & \textbf{Best setting} & \textbf{Portability} & \textbf{Limit} \\
\midrule
VLM & KIMI-K2.5 gives the top NLS evidence score (4.8/5). & Gemma-4-31B \cite{gemmateam2025gemma3} scores 4.7/5; Qwen-122B/397B \cite{qwenteam2025qwen3} are competitive. & Near/sub-12B VLMs add shock, transport, or potential cues in this test. \\
LLM & GLM-5.1: rank-1 exact NLS, broadest true-term recall (9/10). & 6/8 tested LLMs recover exact NLS at rank 1. & Weaker or smaller LLMs need extra iterations or produce cluttered candidate pools here. \\
Pipeline & KIMI-K2.5 VLM + GLM-5.1 LLM is the cleanest tested pair. & Several mid/large alternatives also succeed with Agent~4 filtering. & Evidence for sub-12B backends is limited to easier tested cases. \\
\bottomrule
\end{tabularx}
}
\caption{
LLM/VLM sensitivity analysis on the NLS benchmark: best observed configurations, portability ranges, and practical limits for the VLM and LLM components. Full cue-level rankings refer Appendix.
}\label{tab:llm_vlm_sensitivity}
\end{table}

\subsection{Ablation Study}
We conduct two targeted ablations on the KS benchmark to isolate the contribution of semantic and visual evidence within the pipeline: removing the Phenomenology Extractor (Agent~2) entirely, and removing individual diagnostic panels supplied by the Differential Observer (Agent~1).

\subsubsection{Effect of Agent 2}
This ablation tests whether explicit physical interpretation, instantiated by the Phenomenology Extractor (Agent~2), contributes to recovery on the KS case. Removing this component reduces the pipeline to a data driven functional search without the structured physical abstraction stage. As shown in Table~\ref{tab:agent2-ablation}, the tested configuration then degrades in both structural recovery and coefficient accuracy and does not reach the acceptance threshold. The result supports a contribution from intermediate physical abstraction on KS; broader claims about its necessity require ablations across more equations and random seeds (refer to Appendix).

\begin{table}[htbp]
\centering
{%
\small
\renewcommand{\arraystretch}{1.22}
\setlength{\tabcolsep}{1mm}
\begin{tabular}{>{\raggedright\arraybackslash}p{0.12\textwidth}p{0.31\textwidth}ccccc}
\toprule
\mageTableHead
Configuration & Best selected PDE & $R^2$ & Reward & NRMSE & Confidence & Final status \\
\midrule
\rowcolor{MageGood}
Agent 2 enabled & $u_t = -u u_x - u_{xx} - u_{xxxx}$ & 1.000000 & 0.856864 & 0.000000 & 99.999977\% & \texttt{SUCCESS} \\
\rowcolor{MageBad}
Agent 2 disabled & $u_{tt} = 0.073571 u_{xx} + 0.000938 u_{xxx}$ & 0.252858 & 0.230022 & 0.864374 & 13.56\% & \texttt{EXHAUSTED} \\
\bottomrule
\end{tabular}
}
\caption{End-to-end ablation of Agent 2 on the zero noise KS dataset.}
\label{tab:agent2-ablation}
\end{table}

\subsubsection{Effect of Visual Diagnostic Panels}
We also probe the eight-panel diagnostic evidence produced by the Differential Observer (Agent~1) on the KS case through a leave-one-plot-out ablation, removing each diagnostic in turn while retaining the remaining seven. With the full panel, MAGE accepts the correct KS structure in the first outer iteration with $100\%$ confidence; every single-panel removal instead exhausts the inner loop without acceptance. Removing spectral diagnostics is most damaging, lowering confidence to $6.92\%$ ($E(k)$) and $7.16\%$ ($E(k,t)$), whereas removing local geometric panels (contour, gradient, or extrema views) degrades confidence to a comparatively milder $20$--$55\%$ range. This separation suggests that local panels anchor field geometry while spectral panels supply the modal evidence that fixes derivative order, with both contributing to reliable recovery (refer to Appendix).
\section{Conclusion}
We introduced MAGE, a multimodal agentic framework that reformulates PDE discovery as a confidence driven hypothesis validation loop with explicit separation of perception, physical abstraction, symbolic synthesis, and numerical falsification. On the evaluated cases, MAGE recovers the target structures across the canonical suite, obtains lower aggregate coefficient error than the compared baselines, identifies Laplacian support in two complex geometries, and selects the expected cubic nonlinearity from a laboratory sensor record. The sensitivity and ablation studies further suggest that the workflow can operate with several tested backbones and benefits from intermediate physical abstraction. Together, these findings provide scoped evidence for iterative agentic governing-law discovery rather than a claim of universal generalization.

\paragraph{Limitations}
While MAGE is motivated by a human cognition inspired decomposition of scientific discovery, it relies entirely on off the shelf pretrained VLM and LLM backbones without task specific fine tuning; incorporating domain adapted models within the agentic loop may improve performance. The Equation Arbiter, which performs coefficient estimation via executable regression scripts, remains a bottleneck under noisy observations. This is particularly evident for the PDE-divide example because of its dependence on curvature. Moreover, the experimental evaluation covers one sensor record and a single system class; numerical differentiation amplifies measurement noise, and the held-out record participates in candidate ranking. Evaluation across additional instruments, operating regimes, and genuinely untouched test records is therefore needed before drawing broad conclusions about real-world transfer.

\FloatBarrier

\clearpage
\appendix
\renewcommand{\theequation}{\thesection.\arabic{equation}}
\renewcommand{\thetable}{\thesection.\arabic{table}}
\renewcommand{\thefigure}{\thesection.\arabic{figure}}
\numberwithin{equation}{section}
\numberwithin{table}{section}
\numberwithin{figure}{section}

\section*{Technical Appendix}
\section{Additional details on MAGE}
\label{app:add_det}
In this section, we provide additional details on the proposed agentic framework for equation discovery. In particular, we provide a detailed schematic along with working principle within each agent. 
\subsection{Detailed End-to-End Workflow of MAGE}
\label{sec:appendix_workflow}
The workflow provides a complete execution trace of the MAGE pipeline, illustrating how raw observational data are progressively transformed into validated governing equations through a structured four-agent inference loop. Unlike monolithic discovery systems that directly map measurements to symbolic outputs, MAGE explicitly decomposes discovery into sequential stages of observation, phenomenological abstraction, hypothesis generation, and numerical falsification. This decomposition mirrors the scientific reasoning process of measurement, interpretation, conjecture, and experimental validation.
The pipeline begins with raw field observations supplied as structured arrays together with coordinate metadata. MAGE supports both time-dependent and time-independent settings, including structured spatiotemporal grids, irregular spatial point clouds, and complex geometric domains. These observations are routed first to the Differential Observer.

\paragraph{Agent 1: Differential Observer.}
The Differential Observer converts raw measurements into physics-aware diagnostic representations designed to expose latent structural regularities. This stage computes derivative-informed contour maps, spectral decompositions, and geometric summaries using deterministic numerical operators. For time-dependent fields, the observer generates contour visualizations, derivative fields, spectral density maps, and temporal evolution signatures. For static or geometric domains, it produces contour reconstructions, triangulated surfaces, and projection views adapted to the sampling topology.
This stage performs no semantic reasoning. Its sole purpose is evidentiary transformation: converting numerical measurements into structured visual diagnostics that make transport patterns, oscillatory modes, smoothness transitions, and symmetry properties directly inspectable. The observer additionally emits metadata describing array dimensions, coordinate ranges, discretization resolution, and numerical operators used during construction.
The output of Agent~1 is therefore a multimodal evidence object consisting of visual diagnostics and metadata.

\paragraph{Agent 2: Phenomenology Extractor.}
The evidence object is passed to a vision-language model tasked with identifying latent physical structure. Rather than directly predicting equations, Agent~2 performs constrained phenomenological abstraction.
The system prompt enforces strict reasoning boundaries. The model is prohibited from naming PDEs or writing symbolic expressions. Instead, it must populate a structured JSON schema describing physical cues inferred from the diagnostics.
The extracted evidence includes: (a) inferred dependent and independent variables, (b) data geometry and temporal structure, (c) dominant transport orientation, (d) spectral signatures, (e) smooth-versus-sharp transition behavior, (f) oscillatory or dispersive characteristics, (g) candidate mechanism classes (transport, diffusion, dispersion, instability, source-sink effects), (h) possible coefficient variation signatures.
Each assertion must be explicitly grounded in specific visual panels or observable features. Ambiguity is preserved through admissible \texttt{unclear} entries rather than hallucinated interpretations.
For a representative NLS example, the Extractor identifies nonlinear transport, oscillatory structure, and dispersive wave behavior, producing a JSON evidence file that serves as the semantic prior for symbolic synthesis.
This JSON artifact forms the sole interface between perception and symbolic reasoning.

\paragraph{Agent 3: Governing Law Synthesizer.}
The structured phenomenological evidence is next provided to a large language model specialized for symbolic hypothesis generation.
Conditioned on (a) the Agent~2 evidence JSON, (b) a numerical subsample of the raw data, (c) automated synthesis instructions,
The Synthesizer proposes a slate of candidate governing equations.
Unlike sparse-regression methods, which select terms from a fixed predefined library, the Synthesizer performs unconstrained structural generation. It may propose arbitrary derivative orders, nonlinear interaction terms, mixed operators, or variable-coefficient structures provided they remain consistent with the extracted phenomenology.
Each candidate is emitted in structured canonical form specifying, (a) target derivative term, (b) candidate feature terms, (c) symbolic canonical representation.
The prompt enforces diversity constraints. Candidate sets must span distinct mechanistic families rather than collapsing around local symbolic variants.
In a representative workflow, Agent~3 proposes multiple nonlinear dispersive candidate equations consistent with the VLM evidence.
This stage corresponds to scientific hypothesis formation.

\begin{figure}[t]
    \centering
    \includegraphics[width=\textwidth]{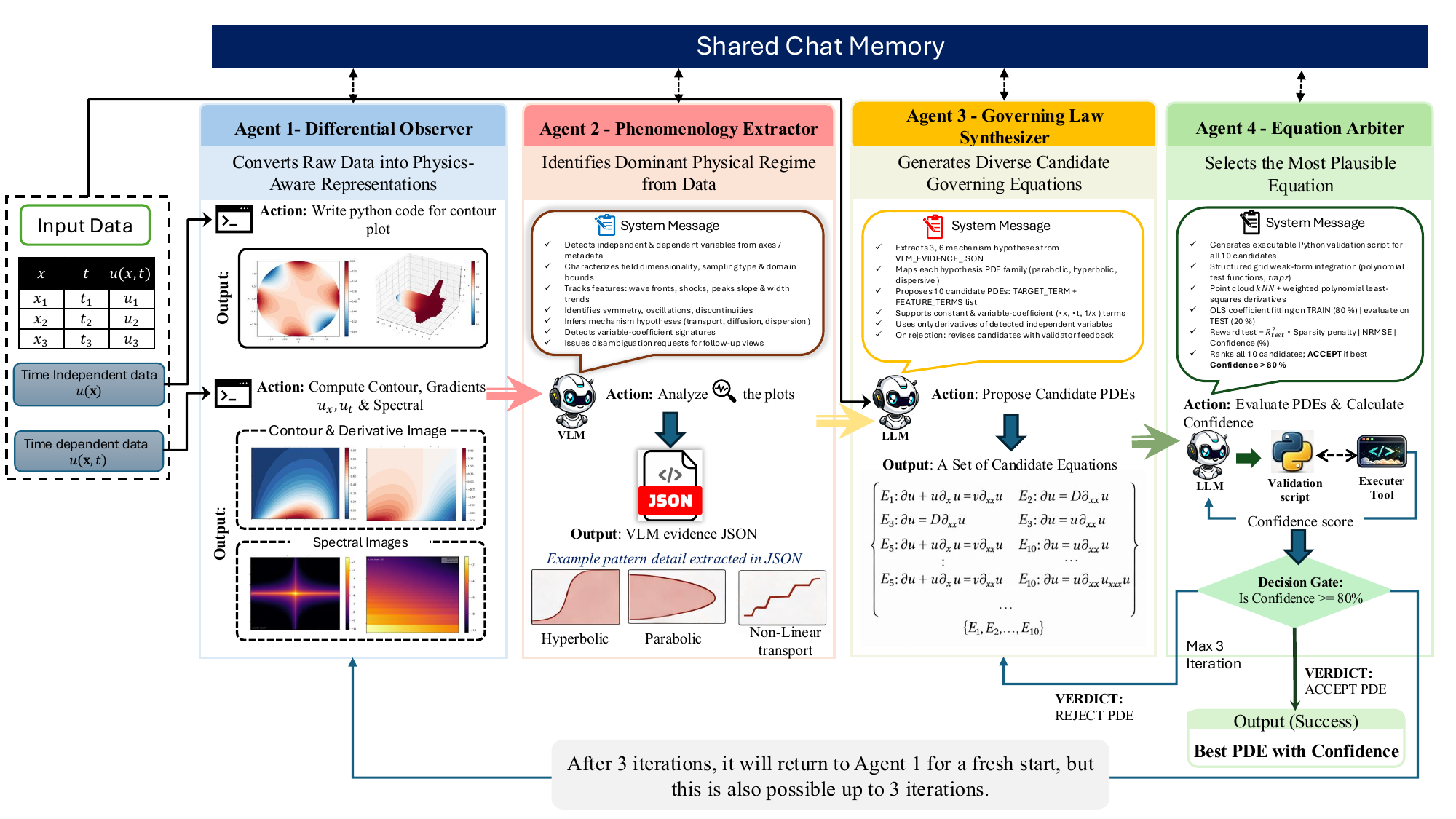}
    \caption{Detailed schematic outlining the working principle of MAGE, the proposed agentic framework for equation discovery.}
    \label{fig:oschematic}
\end{figure}

\paragraph{Agent 4: Equation Arbiter.}
The candidate slate is then passed to the Equation Arbiter for numerical validation.
For each proposed equation, the Arbiter, (a) generates executable Python validation code, (b) computes required weak-form derivatives, (c)  constructs regression targets and feature matrices, (d) estimates coefficients using least-squares regression, (e) evaluates predictive fidelity on held-out data, (f) computes sparsity-aware confidence scores.
The Arbiter operates under strict execution constraints. Candidate equations are evaluated exactly as proposed; structural rewriting, term pruning, or symbolic correction are prohibited. The arbitrar validates using the metric detailed in the main text. 
\paragraph{Closed-loop refinement.}
If no candidate satisfies the confidence threshold, rejection triggers backward propagation of the terminal reward signal. The system then enters an iterative refinement loop. There are two loops,
\textbf{Inner refinement loop.}
Agent~3 receives validator feedback and regenerates a new candidate slate while avoiding previously falsified structural families.
\textbf{Outer restart loop.}
If repeated synthesis attempts fail, the pipeline restarts from Agent~1, forcing a fresh observational interpretation through Agent~2.
As illustrated in Figure~\ref{fig:oschematic}, MAGE permits up to three inner-loop refinements and three outer-loop restarts.
This dual-loop structure is critical. It prevents symbolic search collapse while allowing the system to revise both hypotheses and physical interpretations.
\paragraph{Decision protocol.}
The workflow terminates in one of two states: (a) \textbf{SUCCESS}: a candidate exceeds the confidence threshold and is accepted as the discovered governing equation; (b) \textbf{EXHAUSTED}: all refinement budgets are consumed without numerical validation.
The accepted output consists of (a) final governing equation, (b) fitted coefficients, (c) confidence score.

\subsection{Compute Resources and API-Based Inference}
\label{app:compute_resources}

A primary advantage of the proposed agentic framework is its minimal reliance on local hardware. All experiments were conducted on a standard workstation (6 CPU cores, 32 GB RAM). This local environment was dedicated exclusively to lightweight orchestration tasks: data loading, workflow management, output parsing, validation script execution, and figure generation. Notably, model inference did not necessitate local GPU acceleration or high-performance computing (HPC) clusters. By offloading computationally intensive workloads to remote API endpoints, the framework maintains a negligible local footprint and remains highly accessible to researchers utilizing standard desktop computers.

For inference, the framework leveraged NVIDIA-hosted serverless endpoints via NVIDIA Build/NIM (\url{https://build.nvidia.com/explore/discover}), which provides scalable APIs for deployable foundation models. Specifically, the visual reasoning components were driven by \texttt{moonshotai/kimi-k2.5}~\cite{kimi2025k2}, while language-based reasoning, PDE discovery, and validation tasks utilized \texttt{z-ai/glm-5.1}~\cite{glmteam2024chatglm}---models explicitly optimized for agentic workflows, coding, and long-horizon reasoning. Consequently, the local computational pipeline is streamlined entirely to multi-agent prompting, candidate equation formulation, local code execution for validation, and the routing of self-correction loops. This design choice underscores the framework's accessibility and scalability, enabling researchers to perform state-of-the-art PDE discovery without the need for specialized hardware or extensive local computational resources.

\section{Additional Experimental Analysis }
\label{sec:appendixB}
\subsection{Configuration of Library-Based Methods}
For all library-based discovery methods evaluated in Table~\ref{tab:aggregate} (specifically PDE-FIND, SGA, W-SINDy, Weak-Ident, PINN-SR, and PDE-READ), we employed a consistent, over-complete candidate library to ensure a rigorous comparison. This predefined symbolic library, $\mathcal{F}$, contains the functional forms from which the sparse regression algorithms must select the governing terms:

\begin{quote}
\sloppy\raggedright\footnotesize\ttfamily
"Predefined\_Symbolic\_Library": ["u", "ux", "uxx", "uxxt", "uxt", "ux\^{}2", "uxxxx", "(uux)x", "uxxtt", "(u\^{}4)xx", "u\^{}2", "uxxx", "u\^{}3", "x", "(1/u)xx", "(u\^{}-2*ux)x", "uxxxxx", "sqrt(u)", "sin(u)", "cos(u)", "sinh(u)", "BiLaplace(u)", "Laplace(u)", "x\^{}2", "exp(x)", "sint", "sinx", "(uxx+ux/x)\^{}2", "x\^{}4", "sqrt(x)", "t", "(uux)t", "(uux)xx", "(u\^{}3)xx", "(u(u\^{}2)xx)xx", "|u|\^{}2*u", i, "u\_x/x", "u*u\_x", "u*u\_xx", "u*u\_xxx", "u*u\_xxxx", "u\^{}2*u\_x", "u\^{}2*u\_xx", "u\^{}2*u\_xxx", "u\^{}2*u\_xxxx", "u\^{}3*u\_x", "u\^{}3*u\_xx", "u\^{}3*u\_xxx", "u\^{}3*u\_xxxx"]
\end{quote}

\subsection{Extended VLM--LLM Component Ablation on NLS}
\label{sec:vlm_llm_component_ablation}
This section provides the detailed results underlying the main-text VLM--LLM sensitivity analysis. While the main text reports the key conclusions across foundation-model choices, here we present the full cue-level evidence rankings, candidate-pool audits, and model-wise recovery statistics used to support those findings.
In genral, the VLM--LLM pathway is sequential: the VLM first converts the contour and
spectral diagnostics into phenomenological evidence, and the LLM then turns
that evidence into a finite set of candidate PDEs for validation. A final
coefficient-error heatmap therefore collapses two distinct questions. We
instead audit the NLS case component-wise in
Tables~\ref{tab:nls_component_ablation} and~\ref{tab:nls_llm_ablation}.
The target equation
$i u_t=-0.5u_{xx}-|u|^2u$ is equivalently written by several validators as
$u_t=0.5i\,u_{xx}+i|u|^2u$.

\newcommand{\ablgood}[1]{\textbf{#1}}
\newcommand{\ablmid}[1]{#1}
\newcommand{\ablbad}[1]{\textit{#1}}

\begin{table}[t]
\centering
{%
\footnotesize
\setlength{\tabcolsep}{0pt}
\mageTableRows
\begin{tabular}{@{}>{\raggedright\arraybackslash}m{2.45cm}
                >{\centering\arraybackslash}m{0.9cm}
                >{\centering\arraybackslash}m{0.95cm}
                >{\raggedright\arraybackslash}m{4.25cm}
                >{\raggedright\arraybackslash}m{3.55cm}
                >{\raggedright\arraybackslash}m{4.25cm}@{}}
\toprule
\mageTableHead
\multicolumn{6}{@{}l}{\textbf{Panel A: VLM evidence quality for NLS}} \\
\midrule
\mageTableHead
\textbf{VLM evidence source} & \textbf{Rank} & \textbf{Score}
& \textbf{Correct NLS cues recovered}
& \textbf{Misleading cues}
& \textbf{Component interpretation} \\
\midrule
KIMI-K2.5 & \ablgood{1} & \ablgood{4.8/5}
& Breathing bound-state pattern, closed peak tracks, linear $k$--$\omega$
branches, reversible spectral exchange, entropy oscillations, and no shocks.
& Mild time-varying-coupling cue.
& Richest evidence for smooth conservative NLS dynamics, with a small
overinterpretation risk. \\
\hline
Gemma-4-31B & \ablgood{2} & \ablgood{4.7/5}
& Smooth symmetric peaks, closed tracks, tight $k$--$\omega$ dispersion,
periodic $E(k,t)$ and entropy.
& None material.
& Cleanest evidence for a smooth dispersive breather. \\
\hline
Qwen-3.5-397B & \ablgood{3} & \ablgood{4.5/5}
& Localized breathing, low-$k$ spectral modulation, periodic entropy, explicit
``nonlinear balance'' cue.
& Hints at localized/variable dynamics and clipping.
& Strong NLS description; slight risk of trap/potential terms. \\
\hline
Qwen-3.5-122B & \ablmid{4} & \ablmid{4.0/5}
& High-amplitude temporal events, diagonal dispersion branches, low-$k$ energy
bursts, entropy inversely tied to coherent states.
& Suggests potential wells and time-dependent forcing.
& Good phenomenology, slightly less parsimonious than Qwen-397B. \\
\hline
Nemotron-Nano-12B-VL \cite{nvidia2025nemotron} & \ablmid{5} & \ablmid{3.0/5}
& Recognizes cross-shaped dispersion and periodic spectral entropy.
& Missing a global mechanism summary; emphasizes transient/phase-velocity cues.
& Useful spectral evidence but incomplete as a synthesis prior. \\
\hline
Nemotron-3-Nano-Omni & \ablbad{6} & \ablbad{2.5/5}
& Correct axes, symmetry, and entropy oscillations.
& False sharp/shock cue and only medium confidence on dispersion.
& Mixed evidence; shock language is inconsistent with smooth NLS dynamics. \\
\hline
Microsoft Phi-4 Multimodal \cite{microsoft2025phi4mini} & \ablbad{7} & \ablbad{1.5/5}
& Identifies variables and spectral panels.
& Calls the field steady/sharp, flags shocks, and adds smoothing/transport cues.
& Weakest evidence JSON for NLS; would push the LLM toward non-NLS mechanisms. \\
\bottomrule
\end{tabular}
}
\caption{Component-wise NLS ablation: VLM evidence quality.}
\label{tab:nls_component_ablation}
\end{table}

\begin{table}[t]
\centering
{%
\footnotesize
\setlength{\tabcolsep}{1.5pt}
\mageTableRows
\begin{tabular}{@{}>{\raggedright\arraybackslash}m{2.6cm}
                >{\centering\arraybackslash}m{0.65cm}
                >{\centering\arraybackslash}m{1.0cm}
                >{\centering\arraybackslash}m{1.3cm}
                >{\centering\arraybackslash}m{2.0cm}
                >{\raggedright\arraybackslash}m{7.7cm}@{}}
\toprule
\mageTableHead
\multicolumn{6}{@{}l}{\textbf{Panel B: LLM final-candidate audit with fixed KIMI-K2.5 VLM evidence}} \\
\midrule
\mageTableHead
\textbf{LLM} & \textbf{Iter.} & \textbf{Best PDE}
& \textbf{True-term candidates}
& \textbf{Overcomplete among true-term candidates}
& \textbf{Dominant spurious terms or residual risk} \\
\midrule
DeepSeek-V4-Flash \cite{deepseekai2024v3} & \ablgood{1} & \ablgood{Exact}
& 5/10 & 4/5
& Rank-1 exact; extras add $u_x$, $x^2u$, or $u$ with near-zero coefficients. \\
\hline
DeepSeek-V4-Pro & \ablgood{1} & \ablgood{Exact}
& 4/10 & 3/4
& Rank-1 exact; extras include $u$, $x^2u$, and $|u|^4u$. \\
\hline
GLM-5.1 & \ablgood{1} & \ablgood{Exact}
& 9/10 & 8/9
& Rank-1 exact, but the pool is broad: $u$, $x^2u$, $xu$, $u_{xxxx}$,
$|u|^4u$, $|u|^2u_x$, $|u|^2u_{xx}$, and derivative-NLS variants remain. \\
\hline
Minimax-M2.7 \cite{minimax2025minimax01} & \ablgood{1} & \ablgood{Exact}
& 6/10 & 4/6
& Two clean equivalent forms; extras add $u$, $u_x$, $u_{xxxx}$, or $|u|^4u$. \\
\hline
GPT-OSS \cite{openai2025gptoss} & \ablgood{1} & \ablgood{Exact}
& 8/10 & 6/8
& Broad true-term recall, but many overcomplete variants: $xu_x$, $u_{xxxx}$,
$t^{-1}u$, $|u|^4u$, $|u|^2u_x$, and $|u|^2u_{xx}$. \\
\hline
Nemotron-3-Super & \ablmid{1} & \ablgood{Exact}
& 2/10 & 0/2
& Parsimonious best candidate, but most alternatives use wrong targets such as
$u_{tt}$, $u_x$, or $u_{xt}$. \\
\hline
Qwen-Coder \cite{qwenteam2024qwencoder} & \ablmid{2} & \ablgood{Exact}
& 6/10 & 5/6
& Needs a second iteration; rank-2 is a high-scoring quintic surrogate missing
the cubic term, and true-term variants add $x^2u$, $u$, $xu$, $u_{xxx}$, or
$|u|^2u_x$. \\
\hline
KIMI-Thinking & \ablbad{3} & \ablgood{Exact}
& 9/10 & 8/9
& Highest true-term recall but weakest parsimony: nonlocal, trap, damping,
variable-dispersion, quintic, convective, third-order, and CGLE variants remain
in the final candidate pool. \\
\bottomrule
\end{tabular}
}
\caption{Component-wise NLS ablation: final LLM candidate audit using fixed KIMI-K2.5 visual evidence.}
\label{tab:nls_llm_ablation}
\end{table}

\noindent\textit{Scoring rubric.}
The VLM score uses five one-point cues: correct
variables/grid, smooth localized breathing, dispersive $k$--$\omega$ evidence,
periodic spectral redistribution, and absence of misleading shock, transport,
or variable-coefficient cues. Deductions reflect missing or misleading evidence.
For LLMs, true-term candidates contain both $u_{xx}$ and $|u|^2u$ under an
equivalent NLS target; overcomplete candidates also include extra symbolic
operators, even when fitted near zero.

\noindent\textit{Typography key.}
\textbf{Bold} marks strong or clean evidence and exact rank-1 recovery;
roman type marks useful but mixed results; and \textit{italic} marks weak,
misleading, or heavily overcomplete results.

\paragraph{VLM evidence ranking.}
Among the visual evidence models, KIMI-K2.5 gives the richest NLS description:
it captures localized breathing, closed extrema trajectories, linear
$k$--$\omega$ branches, reversible spectral exchange, entropy oscillations,
and the absence of shocks. Gemma-4-31B is slightly less detailed but cleaner,
because it avoids the mild time-varying-coupling cue present in KIMI-K2.5.
The Qwen VLMs are also strong, but both introduce plausible-looking potential
or variable-coefficient interpretations that are not part of the
constant-coefficient focusing NLS. The Nemotron and Phi-4 models illustrate why
the evidence stage matters: false shock, smoothing, or steady-transport cues
can seed downstream PDE candidates with advection, damping, trap, or higher
order terms even when the data are generated by a smooth dispersive balance.

\paragraph{LLM candidate-set behavior.}
Conditioned on the same KIMI-K2.5 visual evidence, every tested LLM eventually
places an exact NLS form at rank 1. The
meaningful distinction is therefore not the final coefficient error, which is
near zero for all accepted runs, but the parsimony of the final candidate set.
DeepSeek-Flash, DeepSeek-Pro, and Minimax recover the correct equation in one
outer iteration while keeping the overcomplete alternatives relatively simple.
GLM-5.1, GPT-OSS, and KIMI-Thinking achieve broader true-term recall, but most
of those true-term candidates also carry extra operators, indicating a less
selective search prior. Qwen-Coder reaches the exact equation only after a
second iteration and retains a high-scoring quintic surrogate, while
Nemotron-3-Super has a clean best candidate but a candidate pool dominated by
wrong-target alternatives. This ablation supports the component interpretation
of MAGE: VLM quality controls whether the search prior is physically
well-posed, and LLM quality controls whether the final symbolic pool is
parsimonious rather than merely accurate after coefficient fitting.

\subsection{Extended Ablation on Agent 2: Phenomenology Extractor}
\label{app:effect_of_agent2-ablation}

This section provides detailed results supporting the main-text ablation of the Phenomenology Extractor. While the main text highlights the central role of Agent~2 in enabling structured physical abstraction, here we present the full numerical traces, candidate evolution patterns, and failure trajectories observed when this agent is removed.
This ablation evaluates the contribution of Agent 2, the Contour Plot Analyzer, in the proposed agentic PDE discovery system. We compare two runs on the same zero-noise Kuramoto--Sivashinsky (KS) dataset: one with Agent 2 enabled and one with Agent 2 disabled. The dataset is a one-dimensional spatio-temporal field, $u(x,t)$, sampled on a structured grid with $N_x=1024$, $N_t=251$, $\Delta x=0.09817$, and $\Delta t=0.4$; the observed field range is $[-3.02248,3.02248]$. The target governing equation, under the sign convention used by the data and validator, is the KS balance
\begin{equation}
    u_t = -u u_x - u_{xx} - u_{xxxx}.
    \label{eq:ks-ablation-target}
\end{equation}

The ablation shows that Agent 2 is not a cosmetic visualization module. It materially changes the hypothesis-generation stage by translating visual and spectral signatures into a better constrained candidate PDE class. With Agent 2, the system recovers the correct KS equation in the first outer iteration with $R^2=1.000000$, $\mathrm{NRMSE}=0.000000$, $\mathrm{Reward}=0.856864$, and $\mathrm{Confidence}=99.999977\%$. Without Agent 2, the system exhausts its inner loop and selects a low-confidence wave-like surrogate,
\begin{equation}
    u_{tt} = 0.073571 u_{xx} + 0.000938 u_{xxx},
    \label{eq:agent2-disabled-surrogate}
\end{equation}
with $R^2=0.252858$, $\mathrm{NRMSE}=0.864374$, $\mathrm{Reward}=0.230022$, and $\mathrm{Confidence}=13.56\%$. Thus, enabling Agent 2 improves the selected reward by approximately $3.7\times$, increases confidence by about $86.4$ percentage points, and changes the final verdict from \texttt{REJECT\_PDE} to \texttt{ACCEPT\_PDE}.

\subsubsection{Run-Level Outcomes}

Table~\ref{tab:ks_agent2_runs} gives the final best equation from each response log. The with-Agent-2 system consistently identifies the KS equation with coefficients extremely close to the expected unit values, up to the sign convention used by the validator. For example, the first successful run returns
\[
    u_t = -1.000255 u u_x -0.999451 u_{xx} -0.999187 u_{xxxx},
\]
which is numerically indistinguishable from original equation at the level of model structure and coefficient scale. The remaining successful runs similarly recover coefficients near \(1\) in the all-terms-on-left convention or near \(-1\) in the right-hand-side convention.

\begin{table}[t]
\centering
{%
\small
\setlength{\tabcolsep}{3pt}
\mageTableRows
\begin{tabular}{@{}>{\raggedright\arraybackslash}p{2.0cm} c r r >{\raggedright\arraybackslash}p{0.50\textwidth}@{}}
\hline
\mageTableHead
Variant & Run & \(R^2\) & Confidence & Final best PDE \\
\hline
With Agent 2 & 1 & 0.999880 & 98.90 &
\(u_t = -1.000255 u u_x -0.999451 u_{xx} -0.999187 u_{xxxx}\) \\
With Agent 2 & 2 & 0.999056 & 96.93 &
\(u_t = -1.008532 u u_x -1.042205 u_{xx} -1.094190 u_{xxxx}\) \\
With Agent 2 & 3 & 1.000000 & 100.00 &
\(u_t = -1.000000 u u_x -1.000000 u_{xx} -1.000001 u_{xxxx}\) \\
With Agent 2 & 4 & 1.000000 & 100.00 &
\(u_t = -1.0000 u u_x -1.0000 u_{xx} -1.0000 u_{xxxx}\) \\
With Agent 2 & 5 & 0.999999 & 99.90 &
\(u_t +0.999975 u u_x +1.000170 u_{xx} +1.000135 u_{xxxx}=0\) \\
\hline
Without Agent 2 & 1 & 0.252858 & 13.56 &
\(u_{tt}=0.073571 u_{xx}+0.000938 u_{xxx}\) \\
Without Agent 2 & 2 & 0.251817 & 13.50 &
\(u_{tt}=0.091823 u_{xx}+0.015386 u_{xxxx}+0.022255 u_t\) \\
Without Agent 2 & 3 & 0.272681 & 14.72 &
\(u_{tt}=0.030958 u_{xx}+0.013137 u-0.009090 u^3\) \\
Without Agent 2 & 4 & 0.336176 & 18.52 &
\(u_t=-0.351352 u u_x+0.101435 u_{xx}-0.316343 u_x u_{xx}+0.122776 u+0.009089 u^3\) \\
Without Agent 2 & 5 & 0.251277 & 13.47 &
\(u_{tt}=0.046870 u_{xx}+0.015512 u-0.010509 u^3\) \\
\hline
\end{tabular}
}
\caption{Run-level final outcomes from the KS ablation logs.}
\label{tab:ks_agent2_runs}
\end{table}

\subsubsection{Failure Mode Without Agent 2}
\label{app:agent2-failure}

When Agent 2 is disabled, the discovery process begins without the visual mechanism evidence. The initial candidate set is dominated by KdV-like, Burgers-like, and wave-like templates, including $u_t=C_0 u u_x + C_1 u_{xxx}$, $u_t=C_0 u u_x + C_1 u_{xx}$, and $u_{tt}=C_0 u_{xx}$. Crucially, the correct first-order-in-time KS structure containing both $u_{xx}$ and $u_{xxxx}$ together with $u u_x$ is absent. Although $u_{xxxx}$ appears in the disabled run, it is repeatedly attached to second-order time models such as $u_{tt}=C_0 u_{xx} + C_1 u_{xxxx}$, rather than to the correct first-order evolution equation.

This candidate-class misspecification is visible in the validation results. In the first disabled iteration, the selected model is $u_{tt}=0.065263 u_{xx}$, with only $R^2=0.225745$ and $\mathrm{Confidence}=12.01\%$. The best first-order-in-time alternatives perform even worse: for example, $u_t=-0.113519 u u_x + 0.014170 u_{xx}$ has near-zero reward and zero confidence because it lacks the stabilizing fourth-order term. Subsequent retry iterations do not correct the structural error. The final disabled iteration still ranks second-time-derivative models above the first-order KS-like candidates, and the selected final model remains a low-confidence surrogate, $u_{tt}=0.073571 u_{xx} + 0.000938 u_{xxx}$.

The important point is that the disabled system does not fail because the weak-form validator cannot fit derivatives on this dataset. It fails because the candidate generator is not guided toward the correct mechanistic combination of terms. Increasing the number of weak-form supports and retrying candidate generation does not overcome the absence of the key visual prior. The search continues to prefer low-order wave-like explanations because these offer weak partial correlations with the data, even though they do not represent the KS dynamics.

\subsubsection{Interpretation}
\label{app:agent2-interpretation}

The comparison isolates a central benefit of Agent 2: it converts qualitative field morphology into a structured inductive bias over the PDE library. In the KS case, the raw dynamics contain multiple potentially misleading signatures. The diagonal space-time structures can suggest advection or dispersive waves; the oscillatory field can suggest KdV-type dynamics; and low-order temporal correlations can produce spurious $u_{tt}$ candidates with small but nonzero validation scores. Agent 2 resolves this ambiguity by combining several pieces of evidence: nonlinear steepening in derivative panels, bounded high-wavenumber activity, early spectral broadening, and later entropy saturation. The resulting hypothesis is not merely ``include nonlinear and derivative terms''; it is the specific KS balance of nonlinear advection, second-order instability, and fourth-order stabilization.

This also explains the large performance gap. With Agent 2, the correct equation appears in the candidate set before validation, and the weak-form regression only needs to estimate coefficients and rank candidates. Without Agent 2, validation is forced to choose among misspecified candidates. Under such a candidate set, even a reliable validator can only reject or select a poor approximation. Therefore, Agent 2 improves both discovery accuracy and search efficiency: the enabled run terminates successfully in one outer iteration, while the disabled run exhausts its retry budget without reaching the acceptance threshold.

This ablation demonstrates that the Contour Plot Analyser is a necessary component of the agentic PDE discovery system for the KS dataset. It provides mechanistic information that is not captured by scalar metadata alone and steers the discoverer toward the correct first-order-in-time hyperviscous Burgers/KS candidate. Removing Agent 2 causes the system to miss the defining KS term combination, drift toward low-confidence wave-like surrogates, and ultimately reject the discovery. The result supports the broader design principle of the proposed system: visual and spectral reasoning should be treated as an active part of scientific hypothesis generation, not merely as post-hoc interpretability.

For methodological transparency, this is an end-to-end ablation of the agentic pipeline rather than a fixed-validator-only comparison. The validator scripts generated during the two runs use the same weak-form validation philosophy but differ in details such as support placement and candidate lists. This makes the result conservative in the intended sense: the observed failure without Agent 2 is not a small coefficient-estimation discrepancy, but a structural failure to propose the correct PDE class. A fixed-validator replication over identical candidate libraries would be a useful additional diagnostic, but the present evidence already shows that Agent 2 is responsible for introducing the correct KS hypothesis into the discovery process.

\subsection{Finite-Difference Versus Weak-Form Validation}
\label{app:fd-wf-ablation}

This section provides detailed evidence supporting the main-text design choice of weak-form validation. While MAGE can in principle operate with either finite-difference or weak-form derivative estimation within the Equation Arbiter, the results reported here quantify why weak-form validation is preferred, particularly under noisy observations where numerical differentiation stability becomes critical for reliable coefficient estimation and structural acceptance.

We consider the nonlinear Schrodinger (NLS) equation at five additive-noise levels: $0\%$, $1\%$, $2\%$, $5\%$, and $10\%$. The two variants differ only in how the
candidate equations are evaluated from the observed field: the finite-difference (FD) variant fits pointwise differential identities using numerically differentiated data, whereas the weak-form (WF) variant fits integrated identities against compactly supported test functions.

All runs operate on the same space-time grid geometry, with $u\in\mathbb{C}^{512\times 501}$, uniform spacing $\Delta x=0.01953$ and $\Delta t=0.006283$, spatial domain $x\in[-5,4.98047]$, and temporal domain $t\in[0,\pi]$. The target equation identified by the clean FD run and by the weak-form runs is the focusing cubic NLS written in evolution form,
\begin{equation}
    u_t = c_{xx}^{\star}u_{xx} + c_{\mathrm{nl}}^{\star}|u|^2u,
    \qquad
    c_{xx}^{\star}=0.5\,\mathrm{i},\quad
    c_{\mathrm{nl}}^{\star}=\mathrm{i}.
    \label{eq:nls-target-app}
\end{equation} 
Equivalently, this is $i u_t + \frac{1}{2}u_{xx}+|u|^2u=0$ up to the sign convention used in the response logs.

\subsubsection{Main Quantitative Result}
\label{app:fd-wf-main-result}

Table~\ref{tab:fd-wf-summary} summarizes the final validation outcome at each noise level.  The contrast is unambiguous: the weak-form variant accepts the correct NLS model at all five noise levels, while the finite-difference variant accepts the correct NLS model only in the clean $0\%$ case and fails at every nonzero noise level. Averaged over the five runs, WF achieves $99.34\%$ confidence, a mean reward of $0.8991$, and a mean test $R^2$ of $0.9999$, whereas FD achieves $49.94\%$ confidence, a mean reward of $0.4524$, and a mean test $R^2$ of $0.5095$. The minimum WF confidence is $98.72\%$, still far above the acceptance threshold.  In contrast, once any additive noise is introduced, the maximum FD confidence over the nonzero-noise runs is only $51.61\%$.

\begin{table}[t]
    \centering
    {%
    \small
    \setlength{\tabcolsep}{1.5pt}
    \mageTableRows
    \begin{tabular}{@{}c l r r r l r r r@{}}
        \toprule
        \mageTableHead
        Noise & FD verdict & FD $R^2$ & FD reward & FD conf. &
        WF verdict & WF $R^2$ & WF reward & WF conf. \\
        \midrule
        $0\%$  & Accept & $0.999940$  & $0.909637$  & $99.31$ &
        Accept & $1.000000$ & $0.909691$ & $100.00$ \\
        $1\%$  & Reject & $0.439340$  & $0.399663$  & $25.12$ &
        Accept & $0.999836$ & $0.909542$ & $98.72$ \\
        $2\%$  & Reject & $0.668092$  & $0.572464$  & $42.39$ &
        Accept & $0.999909$ & $0.856786$ & $99.05$ \\
        $5\%$  & Reject & $-0.087811$ & $-0.071951$ & $51.61$ &
        Accept & $0.999990$ & $0.909681$ & $99.68$ \\
        $10\%$ & Reject & $0.527865$  & $0.452308$  & $31.29$ &
        Accept & $0.999945$ & $0.909641$ & $99.26$ \\
        \midrule
        Mean & -- & $0.5095$ & $0.4524$ & $49.94$ &
        -- & $0.9999$ & $0.8991$ & $99.34$ \\
        \bottomrule
    \end{tabular}
    }
    \caption{
    Final validation metrics reported by the agentic discovery logs. The acceptance threshold is $80\%$ confidence.  FD reaches the threshold only for the clean $0\%$ case, while WF succeeds at every noise level.
    }
    \label{tab:fd-wf-summary}
\end{table}

The most important qualitative difference is also visible in the accepted equations themselves.  The WF model remains in the cubic-NLS family throughout the full noise sweep, with coefficients close to $(c_{xx}^{\star},c_{\mathrm{nl}}^{\star})=(0.5i,i)$.  In contrast, the FD model is structurally fragile with respect to noise.  At $0\%$ noise, FD finds and accepts the correct two-term NLS equation with high confidence.  Once noise is introduced, however, FD often changes the target from $u_t$ to $u_{tt}$ and selects wave-like or stiff-wave equations, including artificial potentials, fourth derivatives, and high-order nonlinear terms.  This is not a small coefficient perturbation; it is a change in the identified physical model class.

\begin{table}[t]
    \centering
    {%
    \small
    \setlength{\tabcolsep}{2.6pt}
    \mageTableRows
    \begin{tabular}{@{}c >{\raggedright\arraybackslash}p{0.43\textwidth} >{\raggedright\arraybackslash}p{0.43\textwidth}@{}}
        \toprule
        \mageTableHead
        Noise & FD final best PDE & WF final best PDE \\
        \midrule
        $0\%$ &
        $u_t=(0.5i)u_{xx}+i|u|^2u$ &
        $u_t=(0.500000i)u_{xx}+(1.000000i)|u|^2u$ \\
        \addlinespace
        $1\%$ &
        $u_{tt}=5.5753u_{xx}-0.2364x^4u$ &
        $u_t=(-0.001409+0.498327i)u_{xx}
        +(-0.000571+1.000553i)|u|^2u$ \\
        \addlinespace
        $2\%$ &
        $u_t=(0.005518+0.000003i)i\,u_{xx}
        +(0.210816-0.020704i)i\,|u|^2u
        +(3.508637+0.090500i)i\,e^{-x^2}u$ &
        $u_t=(-0.008381+0.503541i)u_{xx}
        +(-0.001953+1.000706i)|u|^2u
        +(-0.000123+0.000055i)u_{xxxx}$ \\
        \addlinespace
        $5\%$ &
        $u_{tt}=(-67.7770+0.1003i)u
        +(18.8292-0.0013i)|u|^2u
        +(18.1959-0.0006i)u_{xx}
        +(0.0054)u_{xxxx}$ &
        $u_t=(-1.14{\times}10^{-4}+0.500661i)u_{xx}
        +(1.55{\times}10^{-4}+0.999476i)|u|^2u$ \\
        \addlinespace
        $10\%$ &
        $u_{tt}=(18.9627+0.0016i)u_{xx}
        +(0.005736+10^{-6}i)u_{xxxx}
        +(10.4994+0.6554i)|u|^2u$ &
        $u_t=(0.000056+0.498982i)u_{xx}
        +(-0.000087+0.997236i)|u|^2u$ \\
        \bottomrule
    \end{tabular}
    }
    \caption{
    Final best PDEs from the response logs. The WF equations remain close to the target NLS form in Eq.~\eqref{eq:nls-target-app}. FD recovers the correct NLS equation in the clean case, but drifts to non-NLS structures when noise is present.
    }
    \label{tab:fd-wf-equations}
\end{table}

\subsubsection{Why Weak-Form Validation Remains Stable?}
\label{app:fd-wf-coefficients}

The superior robustness of weak-form (WF) validation arises from replacing pointwise derivative matching with integral identities. After multiplying a candidate PDE by a smooth test function $\phi$ and integrating over local space--time supports, derivatives are transferred from the noisy field onto the known test function via integration by parts. For the NLS target, this yields the schematic relation
\begin{equation}
  \begin{aligned}
    \int_{\Omega_k} u_t \phi\,dx\,dt
    ={}&c_{xx}\int_{\Omega_k} u_{xx}\phi\,dx\,dt\\
    &{}+c_{\mathrm{nl}}\int_{\Omega_k}|u|^2u\,\phi\,dx\,dt.
  \end{aligned}
\end{equation}
which can be evaluated in a form where differentiation acts on $\phi$ rather than directly on noisy samples of $u$. This simultaneously suppresses high-frequency perturbations through local averaging, reduces sensitivity to discretization noise, and preserves the linear regression structure required for coefficient fitting.

This mathematical smoothing mechanism directly explains the empirical trends observed in Tables~\ref{tab:fd-wf-summary} and~\ref{tab:wf-coef-stability}. Despite evaluating candidate libraries containing linear Schr\"odinger, quintic NLS, drift-augmented variants, and higher-order dispersive alternatives, WF consistently ranks the sparse cubic NLS as the accepted model. Even when noise perturbs the first-pass estimate, refinement remains localized within the correct equation family rather than drifting toward spurious operator classes. At $2\%$ noise, for example, the initial cubic fit is rejected at $77.63\%$ confidence due to coefficient bias, but refinement recovers
$c_{xx}=-0.008381+0.503541i$ and
$c_{\mathrm{nl}}=-0.001953+1.000706i$, raising confidence to $99.05\%$.

Table~\ref{tab:wf-coef-stability} further quantifies this stability. Across all tested noise levels, the average relative error is only $0.52\%$ for the dispersive coefficient $c_{xx}$ and $0.12\%$ for the nonlinear coefficient $c_{\mathrm{nl}}$. The largest deviation occurs at $2\%$ noise, where the accepted model includes a negligible $u_{xxxx}$ correction of magnitude $1.35\times10^{-4}$. This term is best interpreted as a noise-absorbing residual rather than a structural alteration, as the dominant coefficients remain quantitatively aligned with Eq.~\eqref{eq:nls-target-app}.

These results show that WF stability is not merely reflected in high confidence scores. It preserves both the sparse governing structure and physically correct coefficient scales under perturbation, explaining why weak-form validation is the default choice in MAGE.

\begin{figure}[t]
\centering
\includegraphics[width=\textwidth]{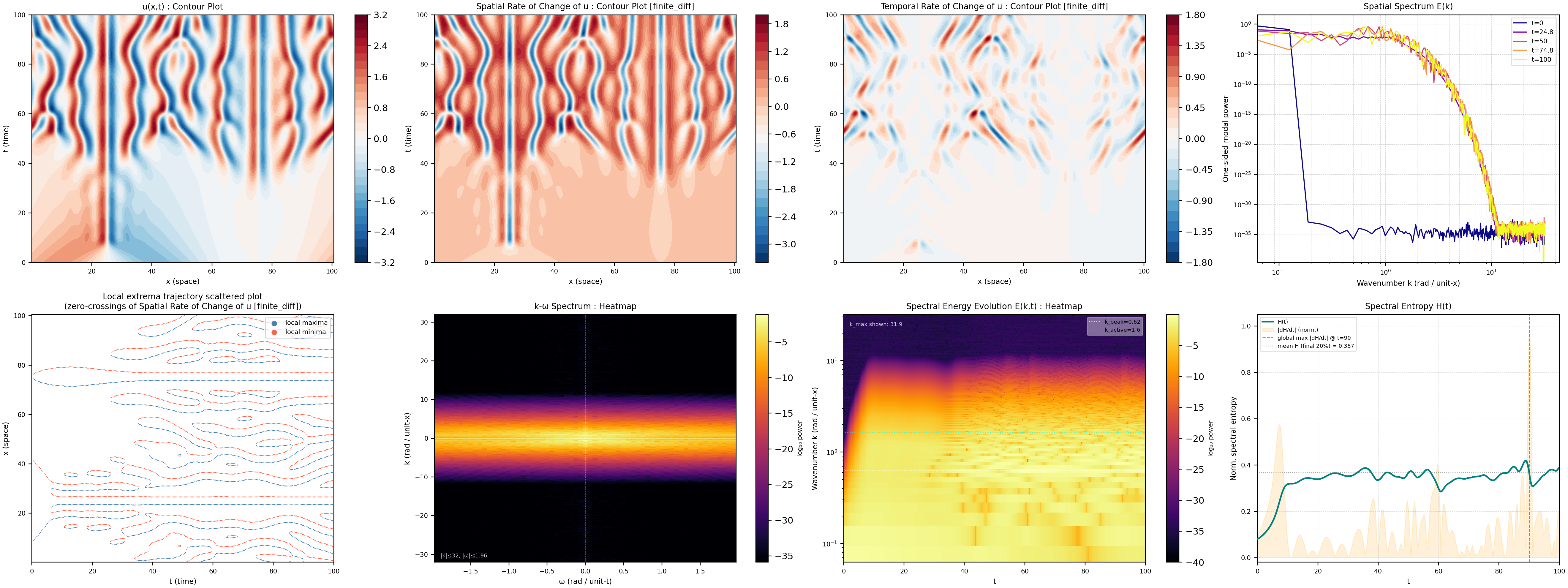}
\caption{%
  Agent~1 diagnostic output for the KS dataset.
  Eight deterministic panels form the evidence object passed to Agent~2.
  \textit{Top:} (1)~field contour $u(x,t)$; (2)~spatial derivative $u_x$; (3)~temporal derivative $u_t$; (4)~spatial spectrum $E(k)$ at five snapshots showing the active band and high-wavenumber rolloff.
  \textit{Bottom:} (5)~local-extrema trajectories; (6)~$k$--$\omega$ dispersion spectrum; (7)~spectral energy evolution $E(k,t)$; (8)~normalized spectral entropy $H(t)$, saturating near $H\!\approx\!0.37$.
}
\label{fig:ks_agent1_panel}
\end{figure}

\begin{table}[t]
    \centering
    {%
    \small
    \setlength{\tabcolsep}{2.3pt}
    \mageTableRows
    \begin{tabular}{@{}c c r c r c@{}}
        \toprule
        \mageTableHead
        Noise & $\hat c_{xx}$ & Rel. err. &
        $\hat c_{\mathrm{nl}}$ & Rel. err. & Extra selected term \\
        \midrule
        $0\%$  & $0.500000i$ & $0.000$ &
        $1.000000i$ & $0.000$ & none \\
        $1\%$  & $-1.409\mathrm{e}{-3}+0.498327i$ & $0.437$ &
        $-5.71\mathrm{e}{-4}+1.000553i$ & $0.079$ & none \\
        $2\%$  & $-8.381\mathrm{e}{-3}+0.503541i$ & $1.820$ &
        $-1.953\mathrm{e}{-3}+1.000706i$ & $0.208$ &
        $|\hat c_{xxxx}|=1.35\mathrm{e}{-4}$ \\
        $5\%$  & $-1.14\mathrm{e}{-4}+0.500661i$ & $0.134$ &
        $1.55\mathrm{e}{-4}+0.999476i$ & $0.055$ & none \\
        $10\%$ & $5.60\mathrm{e}{-5}+0.498982i$ & $0.204$ &
        $-8.70\mathrm{e}{-5}+0.997236i$ & $0.277$ & none \\
        \bottomrule
    \end{tabular}
    }
    \caption{
    Weak-form coefficient stability relative to
    $c_{xx}^{\star}=0.5i$ and $c_{\mathrm{nl}}^{\star}=i$.  Relative errors are
    reported in percent.
    }
    \label{tab:wf-coef-stability}
\end{table}

\subsubsection{Why Finite Differences Fail Under Noise}
\label{app:fd-wf-failure-mode}

The saved logs expose a direct numerical reason for the FD degradation.  The
field amplitude changes only mildly as the nominal noise level increases, but
the finite-difference derivative ranges expand rapidly, especially in time.
Table~\ref{tab:derivative-amplification} reports the metadata ranges from the
FD response logs.  From $0\%$ to $10\%$ noise, the maximum field magnitude
increases by only about $3\%$, whereas the maximum absolute temporal derivative
increases from approximately $4.99$ to $79.33$, a factor of $15.9$.  The maximum
absolute spatial derivative increases from approximately $9.27$ to $23.14$, a
factor of $2.5$.

\begin{table}[t]
    \centering
    {%
    \small
    \mageTableRows
    \begin{tabular}{c c c c c c}
        \toprule
        \mageTableHead
        Noise & $|u|$ range & $u_x$ range & $\max |u_x|$ &
        $u_t$ range & $\max |u_t|$ \\
        \midrule
        $0\%$  & $[0.01599,\,4.00729]$ & $[-9.265,\,9.265]$ & $9.27$ &
        $[-4.994,\,4.993]$ & $4.99$ \\
        $1\%$  & $[0.00991,\,4.01648]$ & $[-9.556,\,9.791]$ & $9.79$ &
        $[-6.865,\,7.452]$ & $7.45$ \\
        $2\%$  & $[0.00300,\,4.01147]$ & $[-10.26,\,10.24]$ & $10.26$ &
        $[-12.72,\,15.44]$ & $15.44$ \\
        $5\%$  & $[0.00039,\,4.09216]$ & $[-11.71,\,13.26]$ & $13.26$ &
        $[-37.24,\,35.81]$ & $37.24$ \\
        $10\%$ & $[0.00030,\,4.12026]$ & $[-23.14,\,16.21]$ & $23.14$ &
        $[-77.46,\,79.33]$ & $79.33$ \\
        \bottomrule
    \end{tabular}
    }
    \caption{
    Noise amplification in finite-difference derivative panels. Even moderate additive noise produces large derivative excursions, particularly for $u_t$.
    }
    \label{tab:derivative-amplification}
\end{table}

This behavior is expected from pointwise differentiation.  If additive noise of
scale $\sigma$ is present in $u$, a $p$th-order derivative estimate amplifies the
noise with a grid-spacing-dependent factor that grows like
$\Delta x^{-p}$ or $\Delta t^{-p}$, up to constants determined by the stencil.
Consequently, the regression target and the candidate library columns become
contaminated by high-variance errors.  In this setting the FD agent can still
recognize NLS-like structure visually, but the subsequent validation step is
forced to explain derivative artifacts.  This explains the observed drift toward
second-time-derivative wave equations, fourth-order spatial derivatives,
external potentials such as $x^4u$ or $e^{-x^2}u$, and high-order nonlinear
terms.  These terms improve noisy pointwise residuals but are inconsistent with
the underlying physics and fail the confidence threshold.

The clean FD run is especially informative in the opposite direction: when
derivative estimates are not contaminated by injected noise, the FD validator
can recover the correct two-term NLS equation with $99.31\%$ confidence.  The
subsequent collapse at $1\%$--$10\%$ noise therefore isolates the failure mode
to noise amplification in the derivative-based validation stage, rather than to
an inability of the agentic system to propose the NLS family.  Under nonzero
noise, this sensitivity becomes a structural-identification failure rather than
merely a coefficient-estimation error.

\subsection{Ablation Study on Agent 1 Outputs}
\label{app:ks_visual_panel_ablation}

We next evaluate how the deterministic visual diagnostics produced by Agent~1 contribute to KS discovery. The full evidence panel contains eight complementary views of the same one-dimensional spatiotemporal field: the primary contour plot $u(x,t)$, spatial derivative contour $u_x(x,t)$, temporal derivative contour $u_t(x,t)$, spatial spectrum $E(k)$, local-extrema trajectories, the $k$--$\omega$ spectrum, spectral energy evolution $E(k,t)$, and normalized spectral entropy $H(t)$. Figure~\ref{fig:ks_agent1_panel} shows all eight panels as produced by Agent~1 on the KS dataset. For each ablation, one plot was removed while the other seven were retained.

As summarized in Table~\ref{tab:ks_visual_panel_ablation}, the full panel produced the accepted KS equation
\[
    u_t = -u u_x - u_{xx} - u_{xxxx},
\]
with $R^2=1.0000$, reward $0.8569$, and confidence $100.00\%$. In contrast, every single-diagnostic removal run failed to meet the $80\%$ acceptance threshold and exhausted the inner loop. With all plots retained, the \texttt{Governing Law Synthesizer} generated one candidate set and the validator accepted a KS-consistent candidate in the same outer iteration. Each ablated condition required three synthesizer rounds, two rejection-router interventions, and ended with \texttt{INNER\_LOOP\_EXHAUSTED}.

\begin{table}[t]
\centering
{%
\small
\setlength{\tabcolsep}{2pt}
\mageTableRows
\begin{tabular}{@{}>{\raggedright\arraybackslash}p{3.0cm} c c c
                    >{\raggedright\arraybackslash}p{0.48\textwidth}@{}}
\toprule
\mageTableHead
\textbf{Removed evidence} & \textbf{Rounds} & \textbf{Status} & \textbf{Confidence} & \textbf{Best validated PDE} \\
\midrule
None (full panel) & 1 & Success & 100.00\% &
$u_t=-u u_x-u_{xx}-u_{xxxx}$ \\
\addlinespace[0.25ex]
$u(x,t)$ contour & 3 & Exhausted & 55.26\% &
\makecell[l]{$u_t=-0.089838u u_x -1.260220u_{xxt}$} \\
\addlinespace[0.25ex]
$H(t)$ entropy & 3 & Exhausted & 49.88\% &
\makecell[l]{$u_{tt}=0.0335u_{xx}-0.8701u_{ttxx}+0.0233u_{xxxx}$} \\
\addlinespace[0.25ex]
$k$--$\omega$ spectrum & 3 & Exhausted & 52.27\% &
\makecell[l]{$u_t=0.0024u^2u_x-0.8778u_{xxt}$} \\
\addlinespace[0.25ex]
Extrema trajectories & 3 & Exhausted & 45.96\% &
\makecell[l]{$u_t=-0.661015u u_x-0.268456u_{xx}+0.010440u^3$} \\
\addlinespace[0.25ex]
$u_x(x,t)$ contour & 3 & Exhausted & 46.77\% &
\makecell[l]{$u_t=-0.848078u_{xxt}-0.027353u u_x+0.000143u_{xxxxx}$} \\
\addlinespace[0.25ex]
$E(k,t)$ evolution & 3 & Exhausted & 7.16\% &
\makecell[l]{$u_t=-0.131729u u_x+0.108460u_{xx}+0.067106u$} \\
\addlinespace[0.25ex]
$E(k)$ spectrum & 3 & Exhausted & 6.92\% &
$u_t=-0.071976u^3u_x-0.008694u_{xx}$ \\
\addlinespace[0.25ex]
$u_t(x,t)$ contour & 3 & Exhausted & 20.05\% &
\makecell[l]{$u_t=-0.000287u^2-0.106348u^3u_x+0.052288u_{xxx}$} \\
\bottomrule
\end{tabular}
}
\caption{Single-diagnostic removal ablation of the Agent~1 visual evidence panel for KS discovery. Only the complete eight-plot panel produced an accepted PDE; every ablated run exhausted the inner loop after three synthesizer rounds.}
\label{tab:ks_visual_panel_ablation}
\end{table}

\paragraph{Mechanistic effect of the full Agent~1 panel.}
With all Agent~1 outputs present, the evidence extractor identifies smooth low-wavenumber initial structure, nonlinear steepening in $u_x$, energy transfer from low to higher wavenumbers, persistent organization near $k\approx0.62$, an active spectral band extending to about $k\approx1.6$, entropy growth followed by saturation near $H\approx0.37$, and bounded oscillatory dynamics. These cues jointly support the KS balance of nonlinear advection $u u_x$, destabilizing second derivative $u_{xx}$, and stabilizing fourth-order dissipation $u_{xxxx}$. In the first synthesizer response, this evidence yields a candidate library spanning nonlinear, second-order, higher-order, and mixed derivative terms, and the KS-consistent candidate
\[
    u_t = C_0 u u_x + C_1 u_{xx} + C_2 u_{xxxx}
\]
appears immediately. The single-diagnostic ablations then reveal which information each panel contributes, and they cleanly separate two roles: local plots anchor field geometry, while spectral plots supply the global modal evidence that pins down derivative orders.

\paragraph{Role of the local panels.}
The local diagnostics, the primary contour $u(x,t)$, the gradient and rate panels $u_x(x,t)$ and $u_t(x,t)$, and the extrema trajectories, encode the field's geometry and propagation. Removing $u(x,t)$ strips away amplitude, phase organization, and coherent space-time morphology, so the remaining cues are no longer anchored to the actual field and the search drifts toward regularized-wave models involving $u_{xxt}$. Removing $u_x$ hides the direct map of gradient sharpening that must be balanced against the stabilizing $u_{xxxx}$ term; without it, steepening can be misread as higher-order dispersion, and the search favors $u_{xxt}$ or $u_{xxxxx}$ over the coupled $\{u u_x,u_{xx},u_{xxxx}\}$ structure. Removing $u_t$ weakens the visual constraint that the dynamics are first-order in time, so KS-like candidates surface only after corrective rounds and the run exhausts before validation. Finally, removing extrema trajectories eliminates path-level evidence of amplitude-dependent propagation, merging, and splitting, leaving the search with insufficient support for the nonlinear transport term and admitting reaction-augmented Burgers/KdV variants as plausible competitors. Across this group, confidence degrades but stays in the $20$--$55\%$ range (Table~\ref{tab:ks_visual_panel_ablation}), reflecting partial loss of geometric grounding rather than collapse of mechanistic identification.

\paragraph{Role of the spectral panels.}
The four spectral diagnostics, $E(k)$, $E(k,t)$, the $k$--$\omega$ spectrum, and $H(t)$, play complementary and non-redundant roles, and together they carry the heaviest mechanistic load. $E(k)$ exposes the active spatial scales and the high-wavenumber rolloff that signals small-scale stabilization rather than mere dispersion or diffusion; removing it collapses final confidence to $6.92\%$. $E(k,t)$ adds the temporal cascade, showing \emph{when} low-wavenumber energy broadens into higher modes and then saturates, and removing it drops confidence to $7.16\%$ and lets the dynamics be misread as generic transport, shock-like steepening, or reaction-diffusion behaviour. The $k$--$\omega$ spectrum supplies the frequency-wavenumber constraint that separates first-order dissipative KS dynamics from regularized wave models; without it, broadband activity is again confused with mixed space-time wave regularization that favours $u_{xxt}$. The entropy $H(t)$ compresses the modal distribution into a single bounded-saturation signal, distinguishing chaotic dissipation from periodic or reversible motion, and its removal nudges the search toward second-order-in-time or regularized-wave candidates. As Table~\ref{tab:ks_visual_panel_ablation} shows, the $6.92\%$ and $7.16\%$ outcomes are far below the $45$--$55\%$ range seen for the local panels, which directly identifies the spectral views as the most discriminative evidence for the fourth-order stabilizing term.

\paragraph{Why spectral plots are central to Agent~1.}
The asymmetry between the two groups explains why we incorporate spectral plots in Agent~1 by default. KdV/Kawahara, Burgers, and BBM candidates can each explain isolated local cues such as oscillations, steepening, or apparent wave regularization, but only the joint spectral signature, simultaneous nonlinear steepening, broadband transfer, bounded entropy saturation, and high-wavenumber suppression, is consistent with the KS balance. Local panels are necessary to anchor the geometry; spectral panels are necessary to suppress these false positives and to identify the correct derivative orders. The correct KS candidate is therefore generated and accepted in the first outer iteration only when the full Agent~1 panel, including all four spectral views, is available.

\section{Running Example of Multimodal Evidence Construction}

\label{sec:appendixA}

\noindent
The user supplies a loaded structured-grid payload, and the system-prompted
\textsc{Differential Observer} returns diagnostic visual evidence for downstream agents.

\begin{tcolorbox}[
    enhanced,
    width=\linewidth,
    colback=orange!20,
    colframe=orange!45!black,
    boxrule=0.45pt,
    arc=2.2ex,
    left=1.25ex,
    right=1.25ex,
    top=1.05ex,
    bottom=1.05ex
]
\small
\hfill\makebox[5.6em][r]{\textbf{\textsc{User}}\hspace{0.65ex}\smash{\raisebox{-0.55em}{\includegraphics[height=3.2em,keepaspectratio]{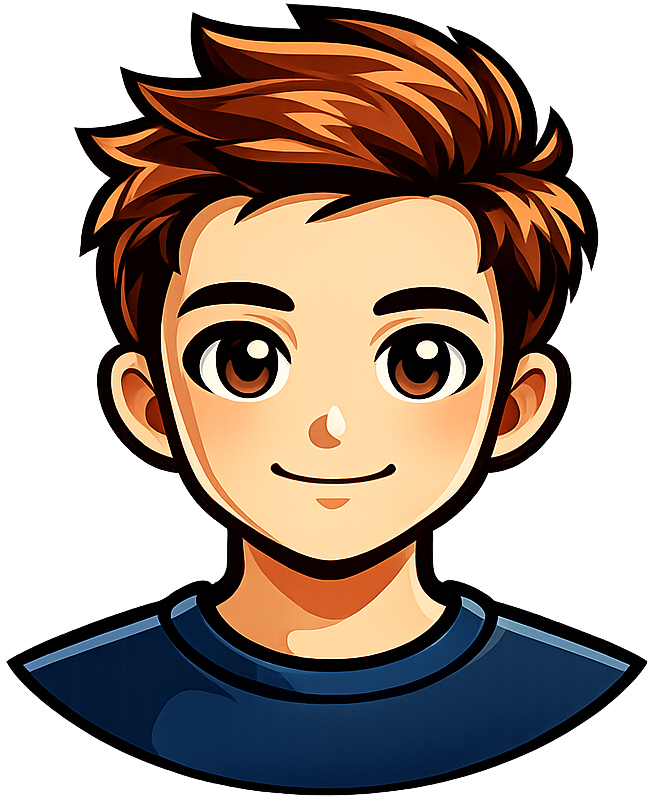}}}}\par
\vspace{0.25ex}
Begin deterministic visualisation panel generation from the currently loaded
\texttt{DATA\_PAYLOAD} for Partial Differential Equation Discovery (PDE).
\end{tcolorbox}

\vspace{0.4ex}

\begin{tcolorbox}[
    enhanced,
    breakable,
    width=\linewidth,
    colback=black!8,
    colframe=black!35,
    boxrule=0.45pt,
    arc=2.2ex,
    left=1.25ex,
    right=1.25ex,
    top=1.05ex,
    bottom=1.05ex,
    before skip=0pt,
    after skip=0pt
]
\small
\makebox[2.55em][l]{\smash{\raisebox{-0.45em}{\includegraphics[height=3.25em,keepaspectratio]{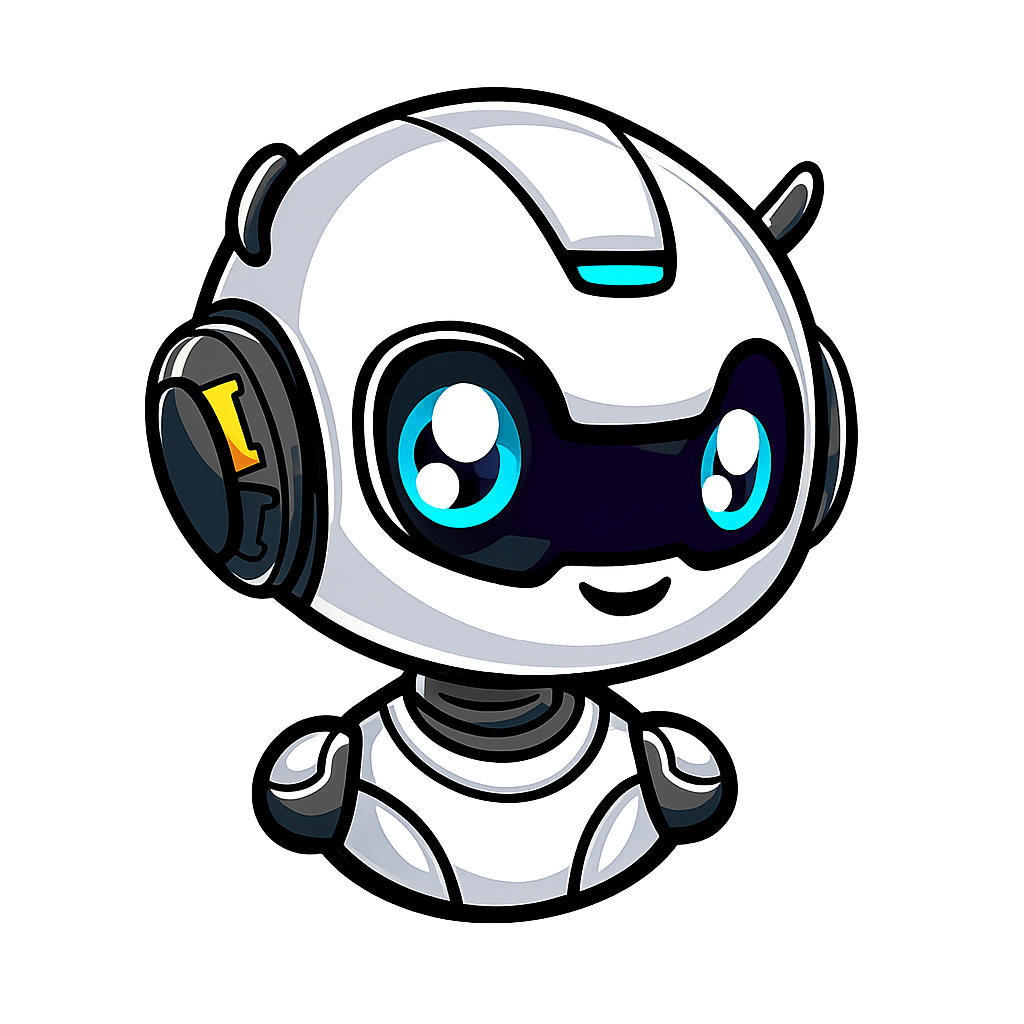}}}}
\hspace{0.6ex}\textbf{\textsc{Agent 1: Differential Observer}}\par
\vspace{0.25ex}
Analyze this panel of figures for Partial Differential Equation (PDE) discovery
and extract qualitative dynamical evidence. \texttt{DATA\_FORMAT=xt\_grid}

\smallskip

{\ttfamily\small
--- DATASET METADATA ---\par
Structured grid summary\par
-----------------------\par
format\hspace{3.4em}: xt\_grid\par
field type\hspace{1.7em}: complex -- panel shows |u|\par
u shape\hspace{2.4em}: (Nx=512, Nt=501)\par
x uniform\hspace{1.7em}: True | dx : 0.01953\par
t uniform\hspace{1.7em}: True | dt : 0.006283\par
\par
Independent Variables :\par
\par
x range\hspace{2.5em}: [-5, 4.98047]\par
t range\hspace{2.5em}: [0, 3.14159]\par
\par
Dependent Variable :\par
\par
u range\hspace{2.5em}: [0.0159882, 4.0073]\par
-------------------\par
Derived Quantities :\par
-------------------\par
All quantities below : derived from u(x,t)\par
-----------------\par
Derivative panels\par
-----------------\par
Spatial Rate of Change of u range\hspace{0.8em}: [-9.265, 9.265]\par
spatial method\hspace{5.5em}: finite\_diff\par
Temporal Rate of Change of u range\hspace{0.5em}: [-4.996, 4.994]\par
temporal method\hspace{5.2em}: finite\_diff\par
-------------\par
Local extrema\par
-------------\par
method : zero-crossings of Spatial Rate of Change of u [finite\_diff]\par
===============\par
Spectral panels\par
===============\par
E(k) -- spatial spectrum\par
------------------------\par
k range\hspace{2.5em}: [0.6283, 160.2] rad/unit-x\par
Nk bins\hspace{2.5em}: 255\par
k\_nyq\hspace{3.1em}: 160.8 rad/unit-x\par
\par
k-omega dispersion\par
------------------\par
k axis displayed : up to 20.36 rad/unit-x\par
omega displayed\hspace{0.7em}: up to 125 rad/unit-t\par
omega\_nyq\hspace{2.1em}: 500 rad/unit-t\par
\par
E(k,t) -- spectral energy evolution\par
-----------------------------------\par
k range\hspace{2.5em}: [0.6283, 160.2] rad/unit-x\par
\par
H(t) -- spectral entropy\par
------------------------\par
Nk bins\hspace{2.5em}: 255  (log(Nk) = 5.5413)\par
normalization\hspace{1em}: H divided by log(Nk), so H in [0, 1]\par
---------------------\par
Pipeline disclosures\par
---------------------\par
FFT windowing\hspace{1.2em}: none\par
detrending\hspace{2.2em}: none\par
DC bin\hspace{3.4em}: excluded from display only (computation unchanged)\par
power floor\hspace{2em}: -40 dB (display only, applied to k-omega and E(k,t) panels)\par
colormaps\hspace{2.4em}: diverging for signed real fields; sequential for magnitudes and log-power\par
--- END METADATA ---\par
}
\end{tcolorbox}

\vspace{0.4ex}

\begin{center}
\begin{tcolorbox}[
    enhanced,
    width=0.86\linewidth,
    colback=white,
    colframe=black!30,
    boxrule=0.35pt,
    arc=1.4ex,
    left=0.85ex,
    right=0.85ex,
    top=0.75ex,
    bottom=0.75ex
]
{\small\textbf{\textsc{Agent Generated Output}}\hfill
\texttt{vlm\_evidence\_panel\_nls.png}}\par
\vspace{0.55ex}
\centering
\includegraphics[
    width=0.98\linewidth,
    height=0.18\textheight,
    keepaspectratio
]{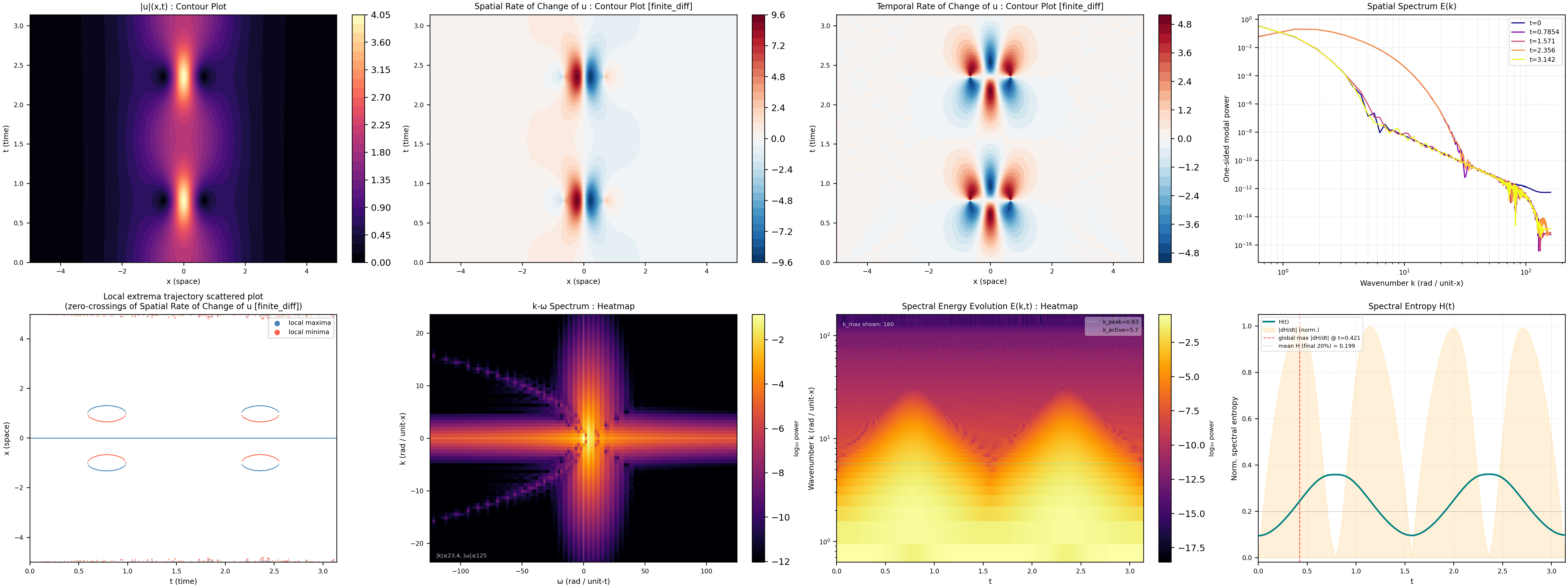}\par
\vspace{0.25ex}
{\small\textit{Visual evidence panel returned by the}
\textsc{Differential Observer}\textit{ for downstream agents.}}
\end{tcolorbox}
\end{center}

\vspace{0.4ex}

\begin{tcolorbox}[
    enhanced,
    width=\linewidth,
    colback=black!8,
    colframe=black!35,
    boxrule=0.45pt,
    arc=2.2ex,
    left=1.25ex,
    right=1.25ex,
    top=1.05ex,
    bottom=1.05ex,
    before skip=0pt,
    after skip=0pt
]
\small
The metadata above is used alongside the panel to produce visual diagnostics
for downstream agents.
\end{tcolorbox}

\vspace{0.9ex}

\begin{tcolorbox}[
    enhanced,
    breakable,
    width=\linewidth,
    colback=orange!20,
    colframe=orange!45!black,
    boxrule=0.45pt,
    arc=2.2ex,
    left=1.25ex,
    right=1.25ex,
    top=1.05ex,
    bottom=1.05ex
]
\small
\hfill\makebox[5.6em][r]{\textbf{\textsc{User}}\hspace{0.65ex}\smash{\raisebox{-0.55em}{\includegraphics[height=3.2em,keepaspectratio]{user-icon.png}}}}\par
\vspace{0.25ex}
Pass the \textsc{Agent 1} output panel
\texttt{vlm\_evidence\_panel\_nls.png}, together with the structured-grid
metadata above, to \textsc{Agent 2}. Use the following system message for the
next agent:

\smallskip

\textbf{System message.} You are a vision-language analyst for PDE discovery.
You must not name specific PDEs or write equations. Extract only evidence and
basic measurements from the multi-panel figure produced by \textsc{Agent 1}.

\smallskip

Infer the independent variables from axis labels and metadata; do not invent
axes; if an axis is absent, mark its fields as \texttt{not\_applicable}; use
\texttt{unclear} only when an existing axis cannot be resolved. Output only
valid JSON. Use \texttt{unclear} rather than guessing, and ground every
evidence field in the figure or metadata panel.

\smallskip

{\small\raggedright
\begin{itemize}
    \item \texttt{variables\_summary}: list PDE independent variables,
    dependent variables, derived axes such as frequency or energy, and a short
    note explaining the classification without introducing new variables.
    \item \texttt{data\_characterization}: report field dimensionality, whether
    time is present, sampling type
    (\texttt{structured\_grid}, \texttt{unstructured\_points},
    \texttt{mesh\_with\_connectivity}, \texttt{particle\_trajectories},
    \texttt{sensor\_network}, \texttt{sliced\_projection}, \texttt{mixed}, or
    \texttt{unclear}), and figure/metadata evidence for that choice.
    \item \texttt{axis\_and\_scaling}: record detected units, linear/log/other
    scales, whether normalization is detected, and a one-sentence note.
    \item \texttt{data\_reliability}: for each variable, give its raw name,
    role guess, uniformity, domain bounds, spacing if available, grounded
    evidence, possible failure modes, colorbar presence, and whether clipping is
    suspected.
    \item \texttt{panels}: for each panel in left-to-right, top-to-bottom order,
    give plot type, raw axis labels, color-axis range, visual character
    (intensity distribution, orientation, sharpness, evolution trend, symmetry,
    spectral falloff, active-band evolution, mode structure, and conjugate-axis
    concentration), a 1--2 sentence grounded comment, and confidence.
    \item \texttt{salient\_observations}: provide 8--14 concrete visual facts;
    mechanism words such as advection, diffusion, dispersion, reaction,
    source, or sink are allowed, but specific PDE names and equations are not.
    \item \texttt{geometry\_and\_motion}: summarize dominant motion direction,
    feature tracks with approximate slopes and track shapes, width trend, and
    amplitude trend.
    \item \texttt{structure\_signatures}: describe smooth/sharp structure,
    discontinuities, oscillations, multi-feature interactions, and symmetry,
    with evidence references to observations or panel comments.
    \item \texttt{mechanism\_possibilities}: list plausible labels such as
    transport, dispersion, smoothing, spreading, source-sink, oscillatory,
    variable-coefficient, coordinate-dependent, anisotropic,
    conservative-flux, instability-growth, saturation-limiting, or mixed, each
    with evidence and confidence.
    \item \texttt{coefficient\_variation\_signatures}: separately record
    position-dependent dynamics, time-dependent dynamics, and other evidence
    for variable coefficients, omitting unsupported entries.
    \item \texttt{disambiguation\_requests}: request 3--6 additional views
    computable from the same data; include time-series diagnostics when time is
    present, otherwise spatial diagnostics.
\end{itemize}
}
\smallskip

\textbf{Input from previous turn.} The attached
\texttt{vlm\_evidence\_panel\_nls.png} contains \(|u|(x,t)\), spatial and
temporal derivative panels, extrema tracks, spatial spectra, a \(k\)--\(\omega\)
diagnostic, spectral-energy evolution, and normalized spectral entropy.
\end{tcolorbox}

\vspace{0.9ex}

\begin{tcolorbox}[
    enhanced,
    breakable,
    width=\linewidth,
    colback=black!8,
    colframe=black!35,
    boxrule=0.45pt,
    arc=2.2ex,
    left=1.25ex,
    right=1.25ex,
    top=1.05ex,
    bottom=1.05ex,
    before skip=0pt,
    after skip=0pt
]
\small
\makebox[2.55em][l]{\smash{\raisebox{-0.45em}{\includegraphics[height=3.25em,keepaspectratio]{AI_Agent_logo.png}}}}
\hspace{0.6ex}\textbf{\textsc{Agent 2: Phenomenology Extractor}}\par
\vspace{0.25ex}
\textbf{\textsc{Response}}\par
\vspace{0.35ex}
{\small\raggedright
\begin{itemize}
    \item \textbf{Variables.} The independent variables are \(x\) and \(t\);
    the dependent field is the displayed magnitude \(|u|\). Derived axes and
    quantities include \(k\), \(\omega\), spatial and temporal rates of change,
    \(E(k)\), \(E(k,t)\), and normalized entropy \(H(t)\).
    \item \textbf{Data characterization.} The data are a two-dimensional
    structured \(x\)--\(t\) grid with time present. Metadata report
    \(N_x=512\), \(N_t=501\), \(dx=0.01953\), \(dt=0.006283\), uniform spatial
    and temporal sampling, and a complex field visualized through \(|u|\).
    \item \textbf{Reliability.} Colorbars are present and no clipping is
    suspected. The plotted field range is approximately \([0,4.05]\), while
    metadata report \(u\in[0.0159882,4.0073]\). Spatial and temporal derivative
    ranges are symmetric and bounded.
    \item \textbf{Panel evidence.} The \(|u|(x,t)\) panel shows two localized,
    high-amplitude events centered at \(x\approx0\), near \(t\approx0.8\) and
    \(t\approx2.4\), with bilateral spatial symmetry. Spatial derivatives form
    antisymmetric dipoles around the center; temporal derivatives form
    oscillatory quadrupole patterns. Extrema tracks remain fixed in space, with
    maxima near \(x\approx0\) and minima near \(x\approx\pm1\).
    \item \textbf{Spectral evidence.} Spatial modal power decays steeply with
    wavenumber, indicating a smooth field. The \(k\)--\(\omega\) panel shows
    strong energy near \(\omega=0\) and diagonal branches with approximate
    slope \(\pm4\)--\(\pm5\). Spectral energy periodically concentrates at low
    \(k\) during high-amplitude phases and redistributes to higher modes during
    collapse phases.
    \item \textbf{Salient observations.} The field has periodic recurrence with
    \(\Delta t\approx1.57\), stationary spatial features, no visible shocks or
    discontinuities, reversible scale exchange, steep high-\(k\) decay, and an
    entropy signal oscillating between roughly \(0.05\) and \(0.35\).
    \item \textbf{Geometry and structure.} Dominant motion direction is
    \texttt{none}; feature tracks are stationary; width is approximately
    constant; amplitude is oscillatory. The structure is smooth, globally
    oscillatory, bilaterally symmetric about \(x=0\), and temporally periodic.
    \item \textbf{Mechanism possibilities.} The evidence supports oscillatory,
    dispersive, nonlinear energy-transfer, and conservative-flux mechanisms.
    There is no grounded evidence for shocks, monotone transport, or irreversible
    smoothing.
    \item \textbf{Disambiguation requests.} Useful same-data diagnostics include
    phase portraits at selected spatial points, time series at \(x=0\) and
    \(x=1\), spatial slices at peak and trough times, complex phase heatmaps,
    integrated energy \(\int |u|^2\,dx\), and group-velocity estimates from
    spectral tracks.
\end{itemize}
}
\end{tcolorbox}

\vspace{0.9ex}

\begin{tcolorbox}[
    enhanced,
    breakable,
    width=\linewidth,
    colback=orange!20,
    colframe=orange!45!black,
    boxrule=0.45pt,
    arc=2.2ex,
    left=1.25ex,
    right=1.25ex,
    top=1.05ex,
    bottom=1.05ex
]
\small
\hfill\makebox[5.6em][r]{\textbf{\textsc{User}}\hspace{0.65ex}\smash{\raisebox{-0.55em}{\includegraphics[height=3.2em,keepaspectratio]{user-icon.png}}}}\par
\vspace{0.25ex}
Pass \textsc{Agent 2}'s \texttt{VLM\_EVIDENCE\_JSON} and the
\texttt{NUMERIC\_SUBSAMPLE} from the loaded \texttt{DATA\_PAYLOAD} to
\textsc{Agent 3}. Use the following system message:

\smallskip

\textbf{System message.} You are a PDE discovery scientist. Propose exactly
10 candidate PDEs that could plausibly generate the observed data. Do not
assume the dataset identity.

\smallskip

{\small\raggedright
\begin{itemize}
    \item \textbf{Inputs.} Use \texttt{VLM\_EVIDENCE\_JSON} as structured
    plot evidence and \texttt{NUMERIC\_SUBSAMPLE} as sampled values of \(u\)
    and its available independent variables.
    \item \textbf{Variable inference.} Determine independent variables from
    the numeric subsample keys or columns first; use visual evidence only if
    needed. Never invent a missing coordinate.
    \item \textbf{Candidate structure.} Each PDE must have exactly one
    physically plausible \texttt{TARGET\_TERM} on the LHS with implicit
    coefficient 1. \texttt{FEATURE\_TERMS} must contain only RHS expressions.
    \item \textbf{Allowed coefficients.} RHS terms may later receive constant
    coefficients \texttt{C0}, \texttt{C1}, \(\ldots\), direct variable factors
    such as \(\texttt{*x}\), inverse factors such as \(\texttt{1/x}\), or
    other independent-variable functions only when supported by evidence.
    \item \textbf{Syntax.} Use derivatives only with respect to detected
    variables and encode them as \texttt{u\_<independent\_variables>}; use
    \texttt{*} for multiplication and \texttt{**} for powers. Do not use
    markdown fences, LaTeX, prose, unsupported terms, or terms that contradict
    the evidence.
    \item \textbf{Search discipline.} Silently extract 3--6 mechanism
    hypotheses from the evidence, map each to plausible PDE families, and
    cover the strongest supported families with parsimonious equations that
    the next agent can validate.
    \item \textbf{Revision rule.} If prior rounds exist, revise using previous
    proposals and validator feedback. Never copy a previous candidate list
    verbatim; reuse a candidate only when the evidence still strongly supports
    it.
    \item \textbf{Output contract.} Return only valid JSON with one key,
    \texttt{CANDIDATE\_PDES}. Each entry must include
    \texttt{EQ\_ID}, \texttt{TARGET\_TERM}, \texttt{FEATURE\_TERMS}, and
    \texttt{CANONICAL\_PDE}. The target term must not appear in
    \texttt{FEATURE\_TERMS}, and feature terms must not include coefficient
    placeholders, labels, prefixes, explanations, or prose.
\end{itemize}
}
\end{tcolorbox}

\vspace{0.4ex}

\begin{tcolorbox}[
    enhanced,
    breakable,
    width=\linewidth,
    colback=black!8,
    colframe=black!35,
    boxrule=0.45pt,
    arc=2.2ex,
    left=1.25ex,
    right=1.25ex,
    top=1.05ex,
    bottom=1.05ex,
    before skip=0pt,
    after skip=0pt
]
\small
\makebox[2.55em][l]{\smash{\raisebox{-0.45em}{\includegraphics[height=3.25em,keepaspectratio]{AI_Agent_logo.png}}}}
\hspace{0.6ex}\textbf{\textsc{Agent 3: Governing Law Synthesizer}}\par
\vspace{0.25ex}
\textbf{\textsc{Response}}\par
\vspace{0.35ex}
{\small\raggedright
According to my analysis, these are the 10 candidate equations for validation.
Format: \textbf{Target} is the left-hand-side term, \textbf{Features} are the
right-hand-side regression terms, and \textbf{PDE} is the full candidate form.\par
\vspace{0.15ex}
\begin{itemize}
    \item \texttt{Equation 1}. \textbf{Target:} \(\texttt{i*u\_t}\);
    \textbf{Features:} \([\texttt{u\_xx},\texttt{|u|**2*u}]\);
    \textbf{PDE:} \(\texttt{i*u\_t = C0*u\_xx + C1*|u|**2*u}\).
    \item \texttt{Equation 2}. \textbf{Target:} \(\texttt{i*u\_t}\);
    \textbf{Features:} \([\texttt{u\_xx},\texttt{|u|**4*u}]\);
    \textbf{PDE:} \(\texttt{i*u\_t = C0*u\_xx + C1*|u|**4*u}\).
    \item \texttt{Equation 3}. \textbf{Target:} \(\texttt{i*u\_t}\);
    \textbf{Features:}
    \([\texttt{u\_xx},\texttt{|u|**2*u},\texttt{|u|**4*u}]\);
    \textbf{PDE:}
    \(\texttt{i*u\_t = C0*u\_xx + C1*|u|**2*u}\)\newline
    \(\texttt{+ C2*|u|**4*u}\).
    \item \texttt{Equation 4}. \textbf{Target:} \(\texttt{i*u\_t}\);
    \textbf{Features:} \([\texttt{u\_xx},\texttt{|u|**2*u},\texttt{u}]\);
    \textbf{PDE:}
    \(\texttt{i*u\_t = C0*u\_xx + C1*|u|**2*u}\)\newline
    \(\texttt{+ C2*u}\).
    \item \texttt{Equation 5}. \textbf{Target:} \(\texttt{i*u\_t}\);
    \textbf{Features:}
    \([\texttt{u\_xx},\texttt{|u|**2*u},\texttt{u\_xxxx}]\);
    \textbf{PDE:}
    \(\texttt{i*u\_t = C0*u\_xx + C1*|u|**2*u}\)\newline
    \(\texttt{+ C2*u\_xxxx}\).
    \item \texttt{Equation 6}. \textbf{Target:} \(\texttt{i*u\_t}\);
    \textbf{Features:}
    \([\texttt{u\_xx},\texttt{|u|**2*u},\texttt{x**2*u}]\);
    \textbf{PDE:}
    \(\texttt{i*u\_t = C0*u\_xx + C1*|u|**2*u}\)\newline
    \(\texttt{+ C2*x**2*u}\).
    \item \texttt{Equation 7}. \textbf{Target:} \(\texttt{i*u\_t}\);
    \textbf{Features:}
    \([\texttt{u\_xx},\texttt{|u|**2*u},\texttt{(|u|**2)\_x*u}]\);
    \textbf{PDE:}
    \(\texttt{i*u\_t = C0*u\_xx + C1*|u|**2*u}\)\newline
    \(\texttt{+ C2*(|u|**2)\_x*u}\).
    \item \texttt{Equation 8}. \textbf{Target:} \(\texttt{i*u\_t}\);
    \textbf{Features:}
    \([\texttt{u\_xx},\texttt{|u|**2*u},\texttt{|u|**2*u\_xx}]\);
    \textbf{PDE:}
    \(\texttt{i*u\_t = C0*u\_xx + C1*|u|**2*u}\)\newline
    \(\texttt{+ C2*|u|**2*u\_xx}\).
    \item \texttt{Equation 9}. \textbf{Target:} \(\texttt{i*u\_t}\);
    \textbf{Features:}
    \([\texttt{u\_xx},\texttt{|u|**2*u},\texttt{|u|**2*u\_x}]\);
    \textbf{PDE:}
    \(\texttt{i*u\_t = C0*u\_xx + C1*|u|**2*u}\)\newline
    \(\texttt{+ C2*|u|**2*u\_x}\).
    \item \texttt{Equation 10}. \textbf{Target:} \(\texttt{i*u\_t}\);
    \textbf{Features:} \([\texttt{u\_xx},\texttt{|u|**2*u},\texttt{x*u}]\);
    \textbf{PDE:}
    \(\texttt{i*u\_t = C0*u\_xx + C1*|u|**2*u}\)\newline
    \(\texttt{+ C2*x*u}\).
\end{itemize}
}
\end{tcolorbox}

\vspace{0.9ex}

\begin{tcolorbox}[
    enhanced,
    breakable,
    width=\linewidth,
    colback=orange!20,
    colframe=orange!45!black,
    boxrule=0.45pt,
    arc=2.2ex,
    left=1.25ex,
    right=1.25ex,
    top=1.05ex,
    bottom=1.05ex
]
\small\raggedright
\hfill\makebox[5.6em][r]{\textbf{\textsc{User}}\hspace{0.65ex}\smash{\raisebox{-0.55em}{\includegraphics[height=3.2em,keepaspectratio]{user-icon.png}}}}\par
\vspace{0.25ex}
Pass \textsc{Agent 2}'s evidence, the 10 \textsc{Agent 3} candidates, and the
saved \texttt{DATA\_PAYLOAD} paths to \textsc{Agent 4}. Use the following
system message:

\smallskip

\textbf{System message.} You are a PDE validation and coefficient-fitting
scientist. Return one complete executable Python script that validates all
10 candidates, fits coefficients on \textsc{Train} only, evaluates on
\textsc{Test} only, ranks the candidates, saves results, and prints the final
verdict.

\smallskip

{\small\raggedright
\begin{itemize}
    \item \textbf{Inputs.} Use \texttt{VLM\_EVIDENCE\_JSON},
    \texttt{PDE\_CANDIDATES} with exactly 10 entries
    (\texttt{EQ\_ID}, \texttt{TARGET\_TERM}, \texttt{FEATURE\_TERMS},
    \texttt{CANONICAL\_PDE}), the saved \texttt{DATA\_PAYLOAD} paths, and the
    data description.
    \item \textbf{No structural edits.} For each candidate,
    \texttt{TARGET\_TERM} defines \(y\) and \texttt{FEATURE\_TERMS} define
    \(\Theta\) columns exactly. Do not add, remove, replace, simplify, prune,
    merge, reorder, factorize, expand, or algebraically rewrite terms. Use
    \texttt{CANONICAL\_PDE} only for reporting.
    \item \textbf{Variable and derivative scope.} Infer available variables
    only from \texttt{DATA\_PAYLOAD}. Never differentiate with respect to a
    missing variable, and compute only derivatives required by each candidate.
    \item \textbf{Structured-grid method.} For \texttt{xt\_grid},
    \texttt{xy\_grid}, \texttt{xyz\_grid}, \texttt{structured\_3D\_grid}, or
    masked structured grids, use weak-form validation as the primary method,
    not pointwise finite differences. For time-independent grids, use only the
    available spatial coordinates and do not invent \(t\) or \(u_t\).
    \item \textbf{Weak test functions.} Use separable compact tests
    \(\phi(s)=(1-s^2)^{2p}\) with \(p=4\); in higher dimensions use products
    over axes. Define normalized support coordinates \(s_i\in[-1,1]\), physical
    coordinates \(r_i=r_{i,c}+L_i s_i\), and \(L_i=H_i d_i\). Compute test
    derivatives analytically with \texttt{numpy.polynomial.polynomial.polyder}.
    \item \textbf{Integration-by-parts rule.} For a derivative term
    \(D^\alpha u\), build the weak contribution as
    \((-1)^{|\alpha|}\prod_i L_i^{1-\alpha_i}\int u\prod_i
    \phi_i^{(\alpha_i)}(s_i)\,ds\). Use trapezoidal integration over normalized
    supports and omit missing axes.
    \item \textbf{Invalid data and support rows.} If \(u\) contains NaNs, use a
    finite-value mask. A support is valid only if all cells in it are finite, or
    if the code uses mathematically consistent masked quadrature. Choose support
    counts and half-widths from grid size, valid-mask density, and term count,
    and justify that choice in one code comment.
    \item \textbf{Point-cloud fallback.} For \texttt{xy\_points} or
    \texttt{xyz\_points}, use kNN plus local weighted polynomial least squares:
    choose degree at least the derivative order, use enough neighbors for the
    monomial basis, extract derivatives from polynomial coefficients, and
    adjust \(k\) or degree if ill-conditioned.
    \item \textbf{Regression system.} For each candidate independently, build
    \(y_j\) from the weak target term and \(\Theta_j\) from one column per RHS
    feature. All candidates must use the global intersection of valid rows.
    Create one deterministic 80/20 split with random seed 0 and reuse it for
    every candidate.
    \item \textbf{Fitting and metrics.} Normalize each
    \(\Theta_{\mathrm{train}}\) column by its train L2 norm, fit by least
    squares, recover physical coefficients, and evaluate only on
    \(\textsc{Test}\). Preserve complex dtype and use magnitude-based residual
    energy. Compute \(R^2\), \(\texttt{NRMSE}=\sqrt{\max(0,1-R^2)}\),
    confidence \(=(1-\texttt{NRMSE})100\), and reward
    \(=R^2\max(0,1-0.3\log_{10}N_{\mathrm{terms}})\).
    \item \textbf{Ranking and files.} Rank by decreasing held-out reward.
    Accept only if \texttt{BEST\_CONFIDENCE > 80}. Do not generate synthetic
    data; raise a clear error if required arrays are missing or incompatible.
    Save the ranked CSV and the script itself as \texttt{pde\_validation\_code.py}.
\end{itemize}
}
\smallskip
{\small\raggedright
\begin{itemize}
    \item \textbf{Output contract.} Return only one fenced Python code block.
    The code must define \texttt{CONFIG}, load the payload, parse candidates
    exactly, infer dataset type, construct valid rows, split once, fit and test
    each candidate, print a ranked table, and finally print
    \texttt{BEST\_R2}, \texttt{BEST\_REWARD}, \texttt{BEST\_NRMSE},
    \texttt{BEST\_CONFIDENCE}, \texttt{BEST\_PDE}, and either
    \texttt{VERDICT=ACCEPT\_PDE} or \texttt{VERDICT=REJECT\_PDE}.
\end{itemize}
}
\end{tcolorbox}

\vspace{0.4ex}

\begin{tcolorbox}[
    enhanced,
    breakable,
    width=\linewidth,
    colback=black!8,
    colframe=black!35,
    boxrule=0.45pt,
    arc=2.2ex,
    left=1.25ex,
    right=1.25ex,
    top=1.05ex,
    bottom=1.05ex,
    before skip=0pt,
    after skip=0pt
]
\small
\makebox[2.55em][l]{\smash{\raisebox{-0.45em}{\includegraphics[height=3.25em,keepaspectratio]{AI_Agent_logo.png}}}}
\hspace{0.6ex}\textbf{\textsc{Agent 4: Equation Arbiter}}\par
\vspace{0.25ex}
\textbf{\textsc{Response}}\par
\vspace{0.35ex}
{\small
Agent~4 returns \texttt{pde\_validation\_code.py}, an executable weak-form
validation script. It treats the 10 candidates only as neutral
\texttt{Equation 1}--\texttt{Equation 10} labels, without using the ID names to
hint at any known PDE family. The script loads \texttt{u.npy}, \texttt{x.npy},
and \texttt{t.npy}; preserves complex dtype; constructs compact supports over
the structured \(x\)--\(t\) grid; evaluates weak-form integrals for
\(\texttt{i*u\_t}\), \(\texttt{u\_xx}\), \(\texttt{u\_xxxx}\), algebraic
nonlinearities, potential terms, and derivative-nonlinear terms; reuses one
deterministic train/test split; fits each candidate by complex least squares;
and saves a ranked CSV. The final printed contract includes the best residual
metrics, fitted PDE, and accept/reject verdict.

\vspace{0.45ex}
\begin{itemize}
    \item \textbf{Validation geometry.} Because the payload is an
    \texttt{xt\_grid}, the Arbiter uses compact weak-form supports rather than
    pointwise finite differences as the primary validation method. It sets
    \(\phi(s)=(1-s^2)^8\), places \(15\times10=150\) support centers, and uses
    support half-widths \(H_x=0.5\), \(H_t=0.3\), large enough for stable
    trapezoidal integration while still giving many more rows than candidate
    terms.
    \item \textbf{Weak-form construction.} Derivatives are moved onto analytic
    test-function derivatives by integration by parts:
    \(\texttt{i*u\_t}\mapsto -i\,u\,\Phi_t\),
    \(\texttt{u\_xx}\mapsto u\,\Phi_{xx}\), and
    \(\texttt{u\_xxxx}\mapsto u\,\Phi_{xxxx}\). Algebraic terms such as
    \(\texttt{|u|**2*u}\), \(\texttt{x*u}\), and \(\texttt{x**2*u}\) are
    integrated directly against \(\Phi\). Mixed derivative-nonlinear terms are
    computed only when required by a candidate.
    \item \textbf{Fair ranking protocol.} All candidate systems use the same
    global valid-support intersection, giving \(150/150\) valid rows. The
    Arbiter then applies one deterministic split
    (\(120\) train, \(30\) test, \(t_{\mathrm{split}}=2.5133\)), normalizes each
    train column by its \(L_2\) norm, fits complex least squares, evaluates only
    on held-out rows, and ranks by held-out reward with a sparsity penalty.
    \item \textbf{Decision rule.} The leading two-term candidate attains
    \(R^2=1.000000\), \(\mathrm{NRMSE}=0.000000\), reward \(0.909691\), and
    confidence \(100.00\%\). Extra-term variants fit the same structure but are
    penalized by parsimony, while the quintic-only alternative falls to
    \(52.15\%\) confidence. Since the best confidence exceeds \(80\%\), the
    Arbiter emits \texttt{VERDICT=ACCEPT\_PDE}.
\end{itemize}
}
\end{tcolorbox}

\vspace{0.45ex}

\begin{tcolorbox}[
    enhanced,
    breakable,
    width=\linewidth,
    colback=white,
    colframe=black!30,
    boxrule=0.45pt,
    borderline west={1.2pt}{0pt}{black!45},
    arc=1.6ex,
    left=1.25ex,
    right=1.25ex,
    top=1.05ex,
    bottom=1.05ex,
    before skip=0pt,
    after skip=0pt
]
\small
\noindent
\begin{tabular}{@{}c@{\hspace{0.95ex}}l@{}}
\includegraphics[height=2.15em,keepaspectratio]{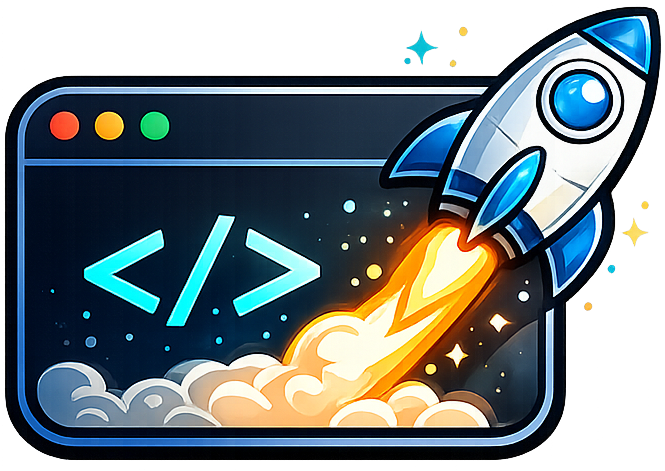}
&
\begin{tabular}{@{}l@{}}
\textbf{\textsc{Code Executor}}\\
\textbf{\textsc{Execution Output}}
\end{tabular}
\end{tabular}\par
\vspace{0.35ex}
{\ttfamily\small
Loading data...\par
u shape=(512, 501), dtype=complex128\par
dx=0.01953,\quad dt=0.006283\par
Supports: 15x10=150,\quad valid supports: 150/150\par
Train=120,\quad Test=30,\quad t\_split=2.5133\par
\par
Rank 1:\quad Equation 1\par
i*u\_t = (-0.500000+0.000000j)*u\_xx\par
\hspace{1em}+ (-1.000000+0.000000j)*|u|**2*u\par
Reward=0.909691,\quad NRMSE=0.000000,\quad Confidence=100.00\par
\par
BEST\_R2=1.000000\par
BEST\_REWARD=0.909691\par
BEST\_NRMSE=0.000000\par
BEST\_CONFIDENCE=100.00\par
BEST\_PDE:\par
i*u\_t = (-0.500000+0.000000j)*u\_xx\par
+ (-1.000000+0.000000j)*|u|**2*u\par
VERDICT=ACCEPT\_PDE\par
TERMINATE\par
SUCCESS in outer iteration 1\par
}
\vspace{0.35ex}
{\small
\textbf{Final discovery summary.}\par
Status: Success; best confidence: \(100.00\%\); threshold:
\(80.0\%\).\par
\textbf{Best PDE:}\par
{\ttfamily\small
i*u\_t = (-0.500000+0.000000j)*u\_xx\par
+ (-1.000000+0.000000j)*|u|**2*u\par
}
}
\end{tcolorbox}

\end{document}